%% file: main.tex
\documentclass{mosi}
\usepackage{helvet}
\usepackage[utf8]{inputenc}
\usepackage{fvextra}

\ifdefined\XeTeXversion
    \microtypesetup{tracking=false}
\fi

\newcommand{\code}[1]{\texttt{\detokenize{#1}}}

\title{\mbox{\raisebox{-0.18\height}{%
    \includegraphics[height=1.35em,trim=72bp 258bp 59bp 232bp,clip]{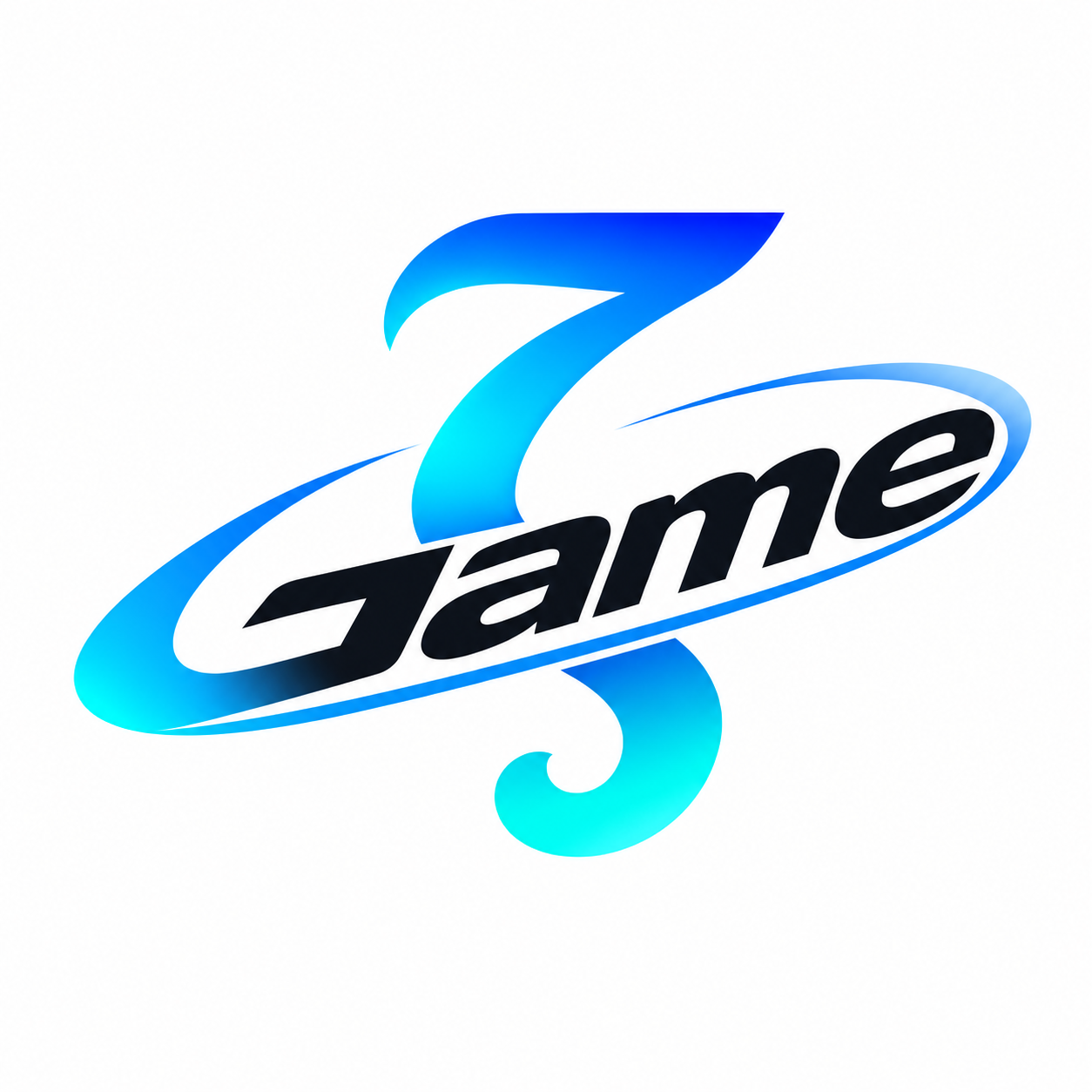}%
}\hspace{0.25em}ZETA:} Towards Real-time AGI in the Frontier AAA Game with Zeroth Embodied Task Agent}
\author{
Jiahao Lu$^{1,2*}$, Shicheng Fang$^{1,2*}$, Jiajun Chen$^{1,2}$, Chuanyuan Tan$^{3}$, Chunbiao Feng$^{1}$, JingJing Gong$^{1}$, Yining Zheng$^{1,2}$, Yuxin Wang$^{1,\dagger}$, Xipeng Qiu$^{1, 2,\dagger}$
\\[2mm]
{\normalfont \normalsize $^{1}$Shanghai Innovation Institute, $^{2}$Fudan University, $^{3}$Suzhou University}\\
{\normalfont \normalsize $^{*}$Equal Contribution}\\
{\normalfont \normalsize $^{\dagger}$Corresponding author:  
wangyuxin@sii.edu.cn, xpqiu@sii.edu.cn}
}

\abstract{
    Frontier AAA games provide a challenging testbed for artificial general intelligence (AGI), requiring agents to master diverse skills, interact with dynamic environments in real time, and pursue complex objectives over extended temporal horizons. While recent game agents have demonstrated promising capabilities in small-scale games and isolated segments of AAA titles, sustained gameplay remains challenging due to highly state-dependent action semantics and the tension between real-time responsiveness and long-horizon reasoning. To address these challenges, we introduce \textbf{ZETA} (Zeroth Embodied Task Agent), a generalist embodied agent powered by our trained vision-language model, \textbf{MOSS-ZETA-preview}. ZETA adopts a dual-system architecture that decouples fast, reactive control (System 1) from long-horizon reasoning and memory management (System 2), allowing the agent to respond to immediate environmental changes while maintaining persistent task awareness. Furthermore, we introduce a dual-path inference infrastructure that addresses KV cache inconsistency arising from history compression, enabling efficient inference over prolonged interactions. ZETA successfully completes 20 hours of main-story gameplay in Grand Theft Auto V (GTA V) and demonstrates promising cross-game generalization to Grand Theft Auto VI (GTA VI) in live-streamed gameplay. We release our dual-system agent framework and inference infrastructure, with the model weights to follow upon the official model release. As a bonus, we release gameplay trajectories collected by ZETA in GTA VI to facilitate future research on world models.
}
\checkdata[Code]{\url{https://github.com/OpenMOSS/MOSS-ZETA}}
\checkdata[Dataset]{\url{https://huggingface.co/datasets/OpenMOSS-Team/GTAVI-preview}}
\checkdata[Project]{\url{https://openmoss.ai/MOSS-ZETA/}}
\begin{document}

\maketitle

\input{sections/01_introduction}

\input{sections/04_agent}
\input{sections/03_data}
\input{sections/05_training}
\input{sections/06_realtime}
\input{sections/06_harness}
\input{sections/10_hardware}
\input{sections/07_experiments}
\input{sections/08_related_work}
\input{sections/09_limitations_conclusion}

\clearpage
\bibliographystyle{unsrtnat}
\bibliography{references}

\clearpage
\appendix
\input{appendices/protocols_and_reproducibility}

\end{document}

%% file: sections/01_introduction.tex
\section{Introduction}
\label{sec:introduction}

Developing agents that can perceive, reason, and act in real-world environments with human-like flexibility is a central goal of artificial general intelligence (AGI). A common approach to developing these capabilities is sim-to-real transfer, in which agents are trained in simulation before being transferred to physical environments~\citep{openai2018dexterity}. Among simulated environments, AAA open-world games offer rich testbeds: tasks such as navigation, vehicle control, and interaction with moving characters demand spatial understanding, temporal reasoning, and effective action execution~\citep{tan2024cradle,chen2026gtagym}. Vision-language models (VLMs) are increasingly used to build agents that translate visual observations and natural-language goals into actions in these environments~\citep{sima2025sima2}. Evaluating these agents therefore requires assessing their ability to achieve task goals through extended interaction with the environment.

\begin{figure}[t]
    \centering
    \includegraphics[width=\linewidth]{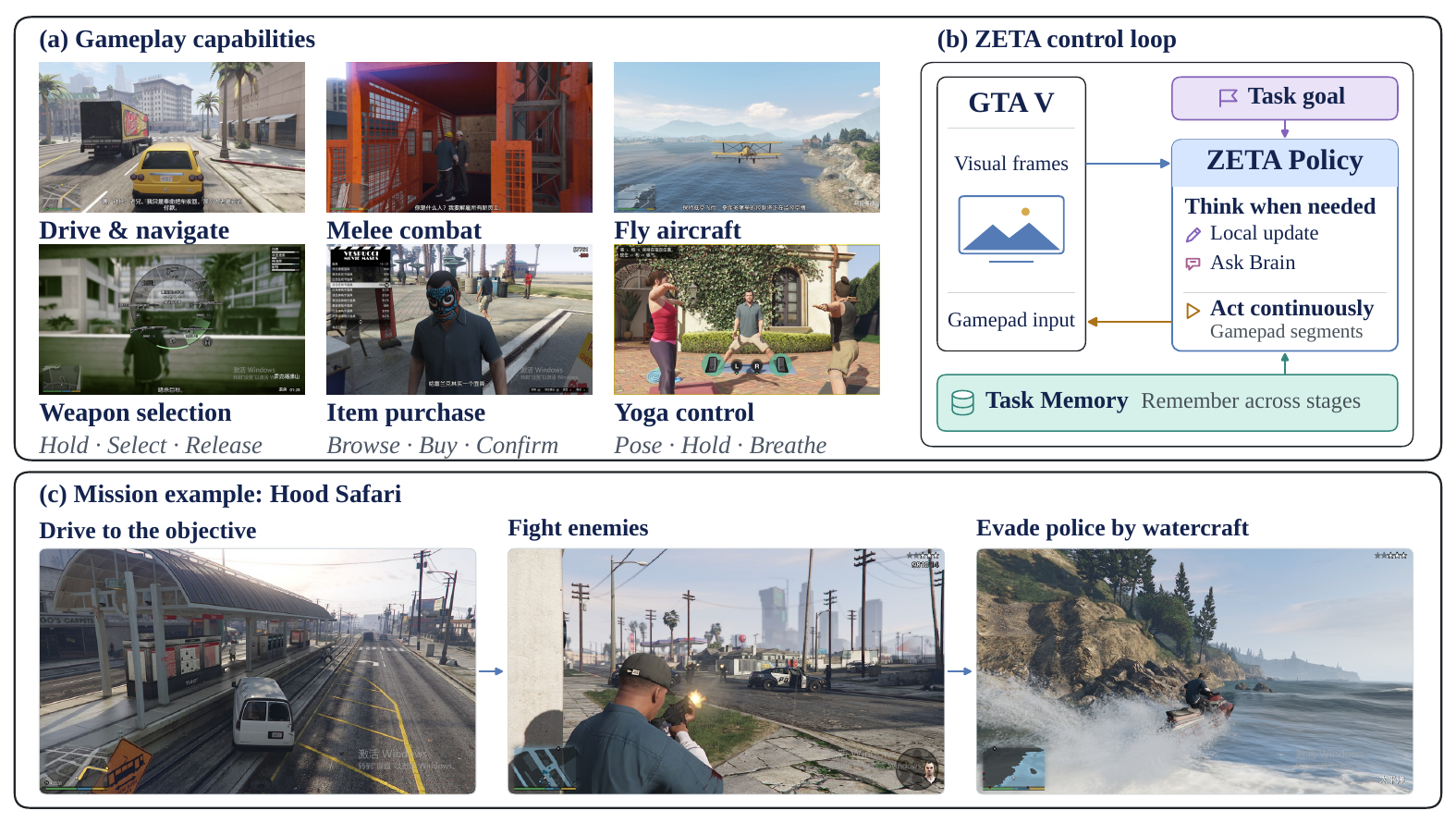}
    \caption{ZETA's gameplay capabilities and control loop. (a) Representative gameplay examples cover driving, melee
    combat, flight, weapon-wheel selection, item purchases, and coordinated yoga inputs. (b) The Policy
    generates short gamepad commands from visual observations, the task goal, and retained memory. It
    updates its local reasoning or asks the Brain to review the current objective as needed. (c) Selected
    moments from the GTA V mission \textit{Hood Safari} show driving to the objective, combat, and police
    evasion by watercraft.}
    \label{fig:zeta-overview}
\end{figure}

Prior work has demonstrated that VLM-based game agents can play small-scale web games, gacha games, and selected segments of AAA titles~\citep{ouyang2026player1,tan2025lumine,tan2024cradle}. However, completing modern AAA games presents a substantially greater challenge, requiring agents to operate over long horizons within a complex and highly state-dependent action space. In Grand Theft Auto V (GTA V), for instance, a player may drive across the city, exit the vehicle, and continue the same mission on foot~\citep{chen2026gtagym}. The semantics of controller inputs vary across gameplay contexts: the same button (e.g., RT) is used to fire a weapon on foot but to accelerate while driving~\citep{rockstarGTAVcontrols}. Moreover, many in-game behaviors require coordinated combinations of multiple controller inputs (e.g., weapon selection), further increasing the complexity of action execution. Figure~\ref{fig:zeta-overview}(a) illustrates this range, from driving, combat, and flight to weapon selection, item purchases, and coordinated yoga inputs. Throughout a mission, an agent must reason across multiple temporal scales, reacting to immediate changes in visual observations while maintaining persistent objectives across gameplay contexts. For example, navigating a turn requires immediate steering adjustments, whereas the destination may remain relevant long after its corresponding on-screen prompt has disappeared.

Real-time control and long-horizon reasoning are both essential yet competing requirements in this setting. Real-time interaction requires agents to process visual observations and execute actions at a high frequency, generating a substantial observation-action history within only a few seconds. Meanwhile, a single mission in an AAA game can span tens of minutes~\citep{tan2024cradle}, requiring the agent to retain and reason over information across much longer temporal horizons. Maintaining the entire interaction history incurs prohibitive computational costs and inference latency~\citep{xiao2023streamingllm}, whereas relying on a short temporal window risks losing critical information necessary for long-horizon task completion~\citep{wang2023jarvis1,li2024optimus1}. This tension poses a fundamental challenge: how can an agent maintain long-term task awareness without compromising real-time responsiveness?

To address these challenges, we present \textbf{ZETA} (Zeroth Embodied Task Agent, where \textit{Zeroth} denotes the stage prior to taking the first action in the physical world), an embodied agent that integrates our trained vision-language model, \textbf{MOSS-ZETA-preview}, with a dual-system (System 1 and System 2) architecture for real-time interaction, long-horizon world exploration, and task execution. We further introduce a dual-path inference infrastructure to address the KV cache inconsistency arising from history compression in conventional approaches, enabling efficient inference over prolonged interactions. ZETA successfully completes 20 hours of main-story gameplay in GTA V and demonstrates promising cross-game generalization to GTA VI in our live-streamed gameplay sessions.

%% file: sections/04_agent.tex
\section{Zeroth Embodied Task Agent}
\label{sec:agent}

\subsection{Task Formulation}
\label{sec:zeta_task}

ZETA receives task instructions and game images and issues gamepad commands.
Before generating each command, the Policy decides whether to continue
the current plan, revise its local reasoning, or ask the Brain to review
the stage objective.

Given the visual input \(o_t\) at time \(t\), bounded observation--response
history \(h_t\), and retained task context \(c_t\), we write the gamepad
segment of the Policy's response as
\begin{equation}
    \hat{A}_t \sim \pi_\theta(\,\cdot\mid o_t,h_t,c_t).
\end{equation}
The segment \(\hat{A}_t\) specifies stick positions and a sequence of button states
for the execution interval \((t+\delta,t+\delta+\Delta]\), where
\(\delta\) is the prediction offset and \(\Delta\) is the segment duration.
The history \(h_t\) captures recent motion and actions, while \(c_t\)
carries the task state across history windows. For example, successive
frames show how the car responds to steering. The full frame shows
nearby traffic, while the minimap from the same capture shows the route
ahead. Task prompts and dialogue read from the screen can also enter the
Policy context when the text is available. Once the prompt naming the
destination leaves the window, the stage instruction still tells the
Policy where to go.

\subsection{ZETA Framework}
\label{sec:zeta_architecture}

\begin{figure}[htbp]
    \centering
    \includegraphics[width=\linewidth]{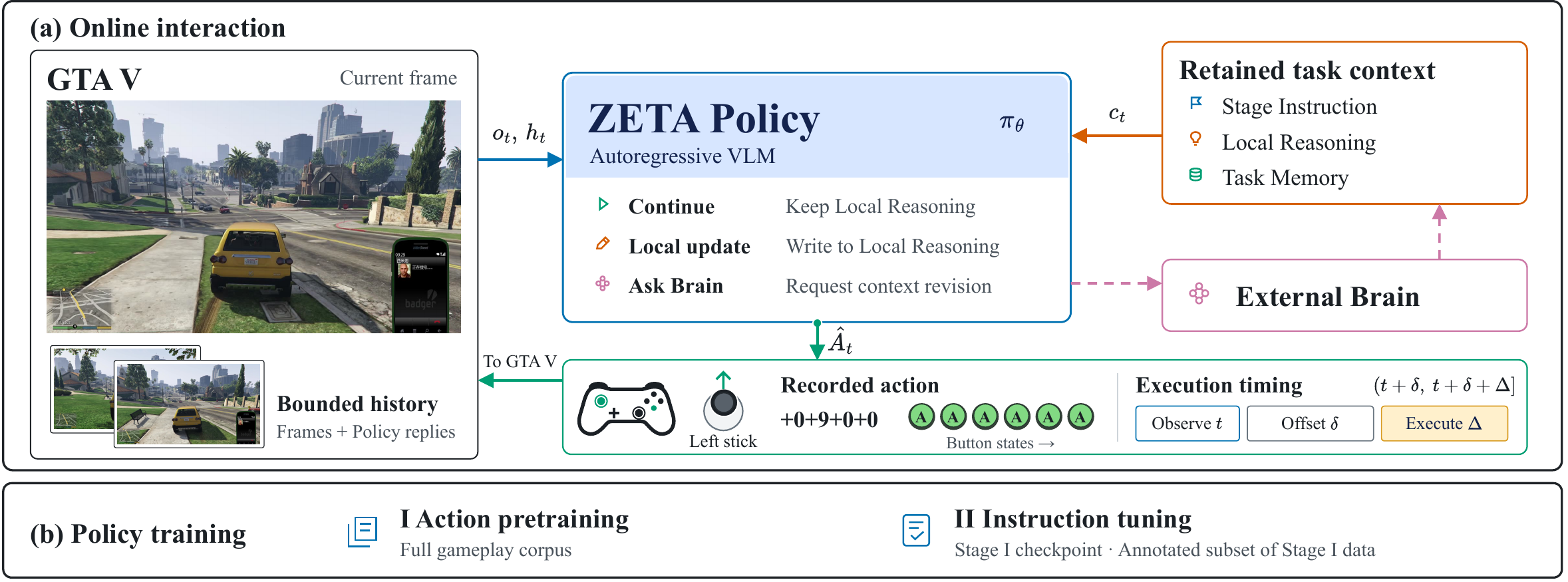}
    \caption{ZETA's online interaction and Policy training. (a) Given the current frame, recent
    interaction history, and retained task context, the Policy generates a gamepad
    segment. It can continue the current plan, revise local reasoning and request a memory
    refresh, or ask the Brain to review the stage objective. The segment covers a 200\,ms
    window beginning 200\,ms after the observation. (b) Action pretraining uses the full
    gameplay corpus. Instruction tuning uses an annotated subset to teach the same Policy
    when to continue or revise its plan.}
    \label{fig:architecture}
\end{figure}

We implement the Policy in Figure~\ref{fig:architecture}(a) as a single
autoregressive model. It generates the update decision and the gamepad
segment in the same response, so the action can depend on the decision
that precedes it. To retain a goal while allowing the
plan to change, we give the three fields of task context distinct roles.
The stage instruction names the active objective, local reasoning
describes how to pursue it, and task memory retains progress and facts.
A blocked road, for example, can prompt a new
route while the destination stays the same.

When the current plan still fits the scene, the Policy emits an empty
think and keeps acting under that plan. Across several such
responses, the Policy can steer or brake in response to each new frame
while following the same route. A nonempty local update
revises the Policy's reasoning and asks the Brain to refresh task memory
under the same stage instruction. When the objective itself needs review,
the Policy requests a full Brain update. The Brain uses the trajectory
to return the stage instruction, task memory, and initial local reasoning
for subsequent decisions. For instance, once a vehicle has been
recovered, a new instruction can direct the agent to deliver it. The
completed retrieval remains part of task memory as the delivery begins.

The execution harness maintains the active context and connects model
responses to physical game inputs. It submits controller segments, records
execution feedback, and activates accepted state updates for subsequent
Policy decisions. Task memory updates can be prepared alongside ongoing
control. A full Brain update pauses play while the Brain reviews the
stage. Its review draws on screenshots, on-screen text, and a record of
which inputs actually executed. After the harness accepts the updated
task state, the Policy resumes from a fresh observation.
To account for the time between observation and execution, we use
\(\delta=\Delta=200\,\mathrm{ms}\). The image at \(t\) conditions
controls for \((t+200\,\mathrm{ms},t+400\,\mathrm{ms}]\). During the
offset, the preceding steering input can remain active while the Policy
computes the next one. We use the same timing in training,
pairing each observation with the recorded controls for that future
interval.

We train the Policy to produce these actions and update decisions in the
two stages shown in Figure~\ref{fig:architecture}(b). Action pretraining
teaches visual control from recorded gameplay. Instruction tuning starts
from this checkpoint and uses a task-directed subset of the same recordings.
The recorded action remains a training target. Annotations at selected
points add targets for revising local reasoning or requesting a stage
review. Continuation examples teach the Policy to act under the retained
plan.

\FloatBarrier

%% file: sections/03_data.tex
\section{Data}
\label{sec:data}

To teach the Policy how to act and when to revise its plan, we draw two
kinds of supervision from human gameplay. We pair observations with the
player's recorded gamepad inputs. For task-directed play, we also annotate
the active instruction, local reasoning, and task memory at the points
where they become relevant.

\subsection{Data Overview}
\label{sec:data_sources}

We collected approximately 4,000 hours of human GTA V gameplay using gamepad
controls, comprising roughly 2,000 hours of free-play without prescribed
task goals and 2,000 hours of task-directed gameplay. Each recording pairs
game video with timestamped controller inputs. We use the entire corpus
for action pretraining.

For instruction tuning, we select approximately 1,200 hours from the
task-directed portion of the same corpus. We supplement these trajectories
with annotations for stage instructions, local reasoning, and task memory.
Table~\ref{tab:corpus_inventory} gives the scale of both datasets.

A stage instruction specifies the current requirement, local reasoning
records the current judgment and plan, and task memory preserves supported
progress and facts. These annotations accompany the visual--action stream
and are retained between updates. Section~\ref{sec:data_annotations}
illustrates these annotations through a shared example and describes their
construction.

\input{tables/corpus_inventory}

\subsection{Visual--Action Alignment}
\label{sec:data_alignment}

We first calibrate the timing of gameplay videos and controller logs,
aligning visual observations and control events on a shared timeline.
For each sampled frame, we use its source-video timestamp, the stick and
button states held before the control window, and events within that window
to reconstruct the complete action label, yielding aligned visual--action
sequences.

We then define the prediction target to account for observation-to-actuation
delay. Given an observation at time \(t\), Policy predicts the complete
gamepad segment for \((t+200\,\mathrm{ms},t+400\,\mathrm{ms}]\), allowing
the preceding segment to execute while the next one is generated.
With observations sampled every 200\,ms, the next observation at
\(t+200\,\mathrm{ms}\) predicts controls for
\((t+400\,\mathrm{ms},t+600\,\mathrm{ms}]\).
The predicted windows are consecutive and non-overlapping because their
200\,ms duration matches the observation interval; the 200\,ms offset
sets their intended execution time after capture. Both sticks and all
button states use this offset, including in historical responses.

This design shares the motivation of real-time chunking (RTC): overlapping
action generation with execution to accommodate inference delay.
RTC adapts existing action-chunking policies by conditioning overlapping
chunks on committed actions~\citep{black2025realtime}. We instead specify
the future action window during training and schedule consecutive segments
for execution, without generating an overlapping action prefix or applying
RTC-style inpainting for faster inference.

\paragraph{Action representation.}
Each 200\,ms action controls the left and right analog sticks and the gamepad
buttons A/B/X/Y, the directional pad, LB/RB, LT/RT, L3/R3, and Menu/View.
We serialize four signed stick coordinates in left-\(x\), left-\(y\),
right-\(x\), right-\(y\) order, followed by six semicolon-separated button
slots inside \code{<a>} tags. Coordinates range from \(-9\) to \(+9\),
with magnitudes at most two mapped to \code{+0}. Each slot represents the
complete pressed-button set for approximately 33.3\,ms; concatenated names
denote simultaneous presses, and an empty slot means no button is pressed.
We repeat one button state over each group of three slots, producing two
ordered 100\,ms blocks. For example,
\begin{verbatim}
<a>+6-3+0+0RT;RT;RT;RTA;RTA;RTA</a>
\end{verbatim}
sets the left stick to \((6,-3)\), centers the right stick, and holds RT
for the first 100\,ms and RT+A for the second. Targets and historical
actions use the same representation, as detailed in Appendix~\ref{app:actions}.

\subsection{Task Annotation}
\label{sec:data_annotations}

We augment task trajectories with stage instructions, local reasoning, and
task memory using DeepSeek-V4-Flash and GLM-5.3-Flash. We first construct and
temporally align the stage instructions, then identify decision points for
joint reasoning and memory annotation. Table~\ref{tab:data_state_roles}
gives a concrete example of these complementary annotations.

\begin{table}[!htbp]
    \centering
    \small
    \begin{tabular}{>{\raggedright\arraybackslash}p{0.17\linewidth}
                    >{\raggedright\arraybackslash}p{0.27\linewidth}
                    >{\raggedright\arraybackslash}p{0.46\linewidth}}
        \toprule
        Annotation & Definition & Example \\
        \midrule
        Stage Instruction & The active task requirement, including its
        object, destination, or completion condition. &
        Drive the car to the dealership. \\
        \addlinespace
        Local Reasoning & An interpretation of the current situation and
        a plan for advancing the active requirement. &
        The car has been recovered; the active goal is to return the yellow
        target SUV to the dealership. After the brief stop at this junction,
        continue along the navigation route. The destination is still some
        distance away and no new obstruction calls for a detour. \\
        \addlinespace
        Task Memory & Confirmed task progress, facts, constraints, and
        historical attempts that may remain useful to later decisions. &
        Franklin was sent by Simeon to retrieve a vehicle from a residence.
        He has entered the yellow target SUV and driven out of the residential
        garage and gate onto a public road. Vehicle retrieval is complete.
        He is now returning to the dealership along the navigation route and
        has not yet arrived. \\
        \bottomrule
    \end{tabular}
    \caption{Task-state annotations in a driving example. The examples are
    abbreviated English renderings of the original annotations associated
    with Figure~\ref{fig:data_example}. Local reasoning interprets the current
    situation and selects how to proceed; task memory retains the preceding
    task context and established progress.}
    \label{tab:data_state_roles}
\end{table}

\subsubsection{Stage Instruction}
\label{sec:data_stages}

Task prompts and dialogue extracted by OCR provide an initial sequence of
stage instructions. We check the objects and goals against the GTA V
mission catalog and gameplay video, then use neighboring frames to locate
each transition. For example, when the player recovers a vehicle, the new
delivery prompt and the state of the vehicle establish when the instruction
to drive to the dealership takes effect.

\subsubsection{Local Reasoning}
\label{sec:data_reasoning}

Within each stage, we examine rolling windows of up to 10 seconds to identify
points that warrant a new judgment, such as selecting a route, choosing an
interaction target, or responding to an obstacle. If such a point is found,
we select it as a non-fallback decision point; otherwise, we use the final
frame of the window as a fallback update point. At both types of point,
we generate local reasoning and updated task memory from the active
instruction, visual/OCR evidence, and prior state. Local reasoning may
reference up to three seconds of subsequent gameplay to clarify the current
judgment and plan. Task memory is restricted to facts established in the
selected segment and preceding state by the selected node.
Appendix~\ref{app:annotations} details these annotation constraints.
The next search starts after the chosen node.

At non-fallback decision points, the new local reasoning is supervised as a
Policy response. At fallback update points, the generated local reasoning
serves only as subsequent input context and receives no output supervision.
Task memory is always supplied as input and is never a Policy output target.
It summarizes trajectory context that is not fully available in the Policy's
bounded visual history.

\subsubsection{Task Memory}
\label{sec:data_memory}

At the same annotation points, we combine the previous task memory with
new visual/OCR evidence to produce a complete updated memory. It preserves
confirmed progress and useful facts, adds newly established information,
and corrects or removes obsolete content. Proposed plans belong to local
reasoning; task memory retains supported knowledge that may remain useful
across stage transitions and visual-history eviction.

\subsection{Sample Construction and Augmentation}
\label{sec:data_examples}

\paragraph{Sample assembly.}
Action-pretraining samples pair temporally ordered game frames with recorded
controller actions, using the alignment and representation in
Section~\ref{sec:data_alignment}. Instruction-tuning samples add the active
stage instruction, retained local reasoning, and task memory constructed in
Section~\ref{sec:data_annotations}. Annotations take effect at their aligned
times and are carried forward until updated. For instruction tuning, each
observation includes the full game frame, a same-frame minimap, and available
OCR; preceding observation--response pairs provide recent interaction history.

The same local-reasoning annotation can serve different roles: retained or
fallback reasoning supplies context, whereas reasoning at a non-fallback
decision point provides an output target. Task memory supplies context throughout.
Section~\ref{sec:training_details} specifies the context windows and
supervision masks; Appendix~\ref{app:training_samples} details message
organization and prompt fields.

\paragraph{Response format.}
Instruction-tuning responses combine an update decision with a recorded
action. The following patterns show continuation, a local reasoning update,
and a Brain request, respectively:
\begin{verbatim}
<think></think><a>{action}</a>
<think>{reasoning}</think><a>{action}</a>
<think>{reasoning}<|call_brain|>{payload}</think><a>{action}</a>
\end{verbatim}
Braced names are placeholders. An empty think retains the preceding local
reasoning; a nonempty local update places the selected reasoning annotation
from Section~\ref{sec:data_reasoning} inside \code{<think>} tags. At annotated
stage-reassessment points, a pre-call judgment is followed by
\code{<|call_brain|>}. In these training examples, the following JSON payload
contains \code{instruction}, the next stage instruction, and
\code{small_think}, its initial local reasoning, connecting the stage and
reasoning annotations in Section~\ref{sec:data_annotations}. The updated
context becomes available from the following round. Task memory remains an
input field supplied externally. All three forms end with the recorded
action inside \code{<a>} tags; its internal format follows
Section~\ref{sec:data_alignment}.

\begin{figure}[!t]
    \centering
    \includegraphics[width=\linewidth]{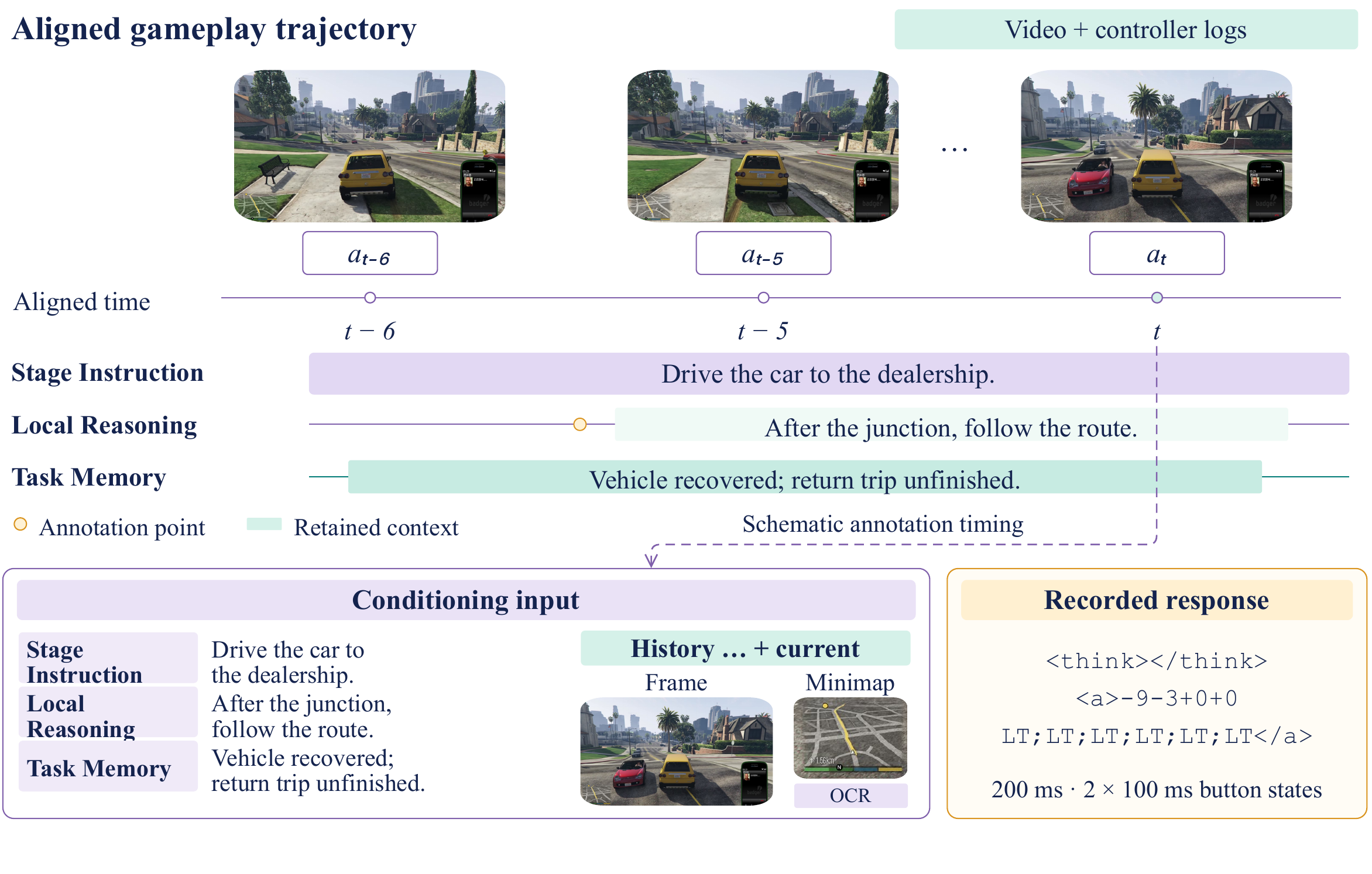}
    \caption{From a recorded trajectory to an instruction-tuning sample. Frames, controller inputs,
    and task annotations share a timeline. In this example, memory records that the car
    has been recovered, and the stage instruction directs the player back to the
    dealership. At the marked observation, the Policy receives the active context, recent
    history, current frame and minimap, and available OCR. The recorded response continues
    the plan with an empty think and a 200\,ms gamepad action. Annotation timing is
    schematic, and the text is abbreviated from the driving example in
    Table~\ref{tab:data_state_roles}.}
    \label{fig:data_example}
\end{figure}

\paragraph{Reading the example.}
In Figure~\ref{fig:data_example}, the player is driving the recovered car
back to the dealership. The stage instruction gives the destination, and
task memory records that the player has already retrieved the car.
At this point the player continues along the route, so the sample pairs
an empty think with the recorded gamepad input for the 200\,ms target window.

\paragraph{Augmentation for target capabilities.}
Closed-loop failure cases and review of capability coverage guide augmentation
of selected on-foot and vehicle intervals. We construct additional
observation--action streams with a 100\,ms sampling-phase offset, aligning
images and controller labels at the new observation times using
Section~\ref{sec:data_alignment}. This supplies more temporal views of the
selected behavior while retaining its stage instructions, local reasoning,
and task memory.

During evaluation, we observed a pattern of consecutive empty actions,
with both sticks centered and all buttons released. To train recovery from
this pattern, we prepend five repeated observations paired with empty
actions to the original expert continuation. The prefix serves as input history. The expert continuation
supplies the action targets under the original task context.

%% file: tables/corpus_inventory.tex
\begin{table}[!htbp]
    \centering
    \small
    \begin{tabular}{lr}
        \toprule
        Training data & Hours (approx.) \\
        \midrule
        Action pretraining & 4,000 \\
        \quad Free-play & 2,000 \\
        \quad Task-directed & 2,000 \\
        \midrule
        Instruction tuning & 1,200 \\
        \bottomrule
    \end{tabular}
    \caption{Approximate scale of the gameplay data. Instruction tuning uses
    an annotated task-directed subset of the action-pretraining corpus.
    Hours count original recordings before augmentation.}
    \label{tab:corpus_inventory}
\end{table}

%% file: sections/05_training.tex
\section{Training}
\label{sec:training}

\begin{figure}[!t]
    \centering
    \includegraphics[width=\linewidth]{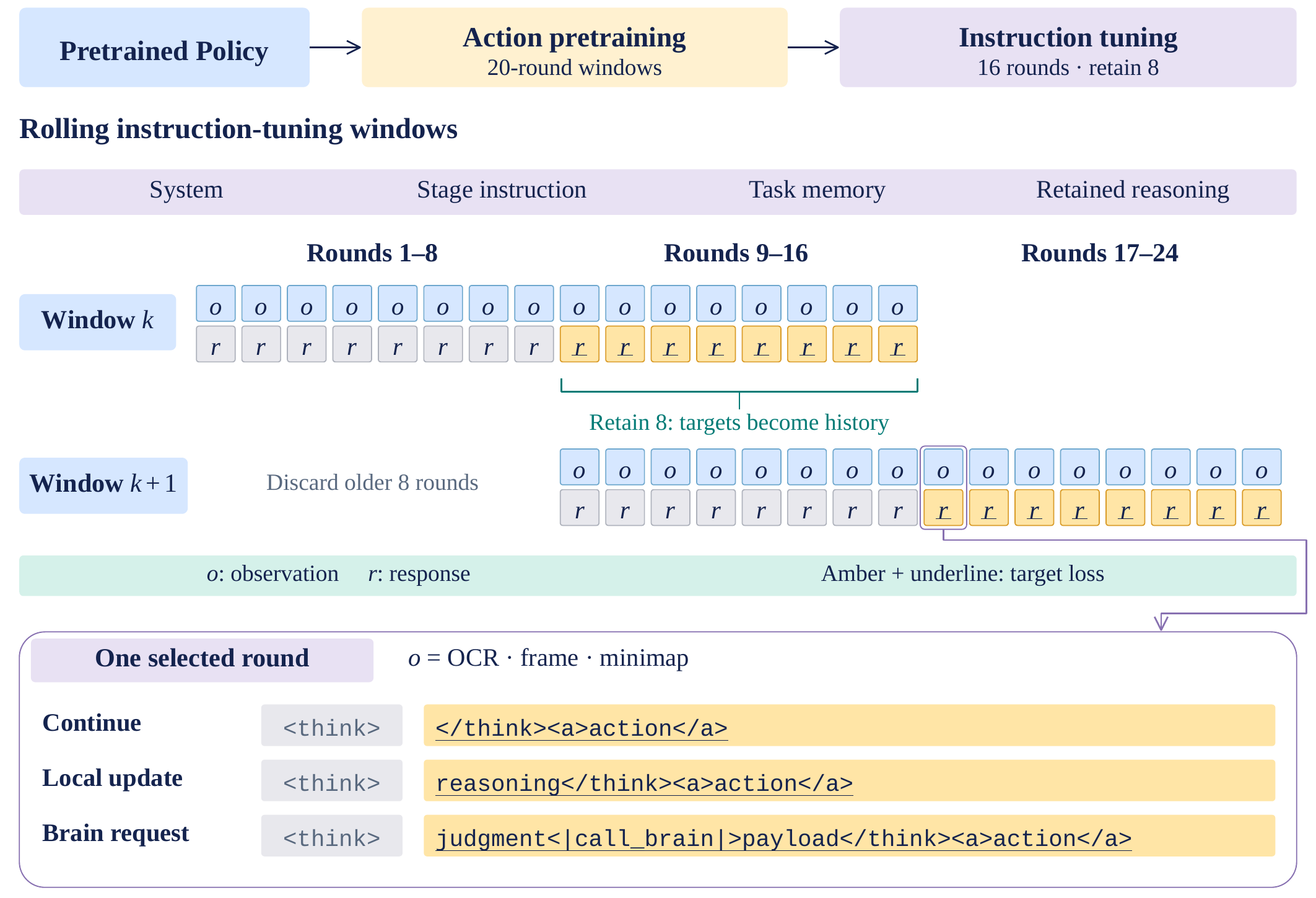}
    \caption{Policy training and rolling-window supervision.
    Action pretraining uses 20-round windows; instruction tuning uses
    16-round windows with eight retained rounds. Each paired \(o/r\) column
    is one observation--response round. Rounds 9--16 align across consecutive
    windows: their selected responses change from supervised targets
    (amber, underlined) to input-only history (gray), while new rounds extend
    the window. The inset expands one round and shows three alternative
    response continuations. The opening \code{<think>} is input; each amber
    continuation, including its closing tag and action, is supervised. Reasoning, payload, and
    action strings are placeholders. The illustrated windows select all new
    responses; task context follows its aligned update times.}
    \label{fig:training_stages}
\end{figure}

We use the gameplay data from Section~\ref{sec:data} to train a single
autoregressive Policy in two stages. We first pretrain it to predict
gamepad controls from recorded play. We then use the annotated
trajectories to teach the same Policy when to continue a plan, revise
local reasoning, or request a Brain update. Figure~\ref{fig:training_stages}
shows how interaction rounds form training windows and which response
spans receive supervision.

\subsection{Action Pretraining}
\label{sec:training_action}

Policy combines a visual encoder, multimodal projection, and an autoregressive
language model. Initialization from pretrained multimodal weights supplies
visual and language representations, which action pretraining adapts to
gamepad control. Controller coordinates and button states are generated as
tokens in the response stream.

The first stage uses the visual--action data in
Section~\ref{sec:data_examples}. Its prompt specifies the controller response
format and asks the Policy to predict an action from visual interaction
history. Consecutive observations and preceding actions supply motion and
control context; each response is supervised against the associated recorded
controller interval. This stage establishes the visual control prior used
by subsequent task-conditioned training.

\subsection{Instruction Tuning}
\label{sec:training_reasoning}

Instruction tuning initializes from the action-pretrained Policy with a fresh
optimizer and learning-rate schedule. We fine-tune the model parameters using
the task-conditioned samples in Section~\ref{sec:data_examples}. Stage
instructions, local reasoning, and task memory allow action prediction to
depend on task objectives and progress beyond the visible interaction history.

The recorded gamepad action is a target in each selected response.
Consider the return trip to the dealership in Figure~\ref{fig:data_example}.
The Policy reads from memory that the car has already been recovered.
At the step shown, it learns to emit an empty think to continue the plan
and to predict the driver's next gamepad action. At a local decision
point, the target also includes new reasoning. When a stage needs
reassessment, the Policy learns to generate a Brain request with the
proposed next-state fields. Task memory and retained or fallback reasoning
supply input context throughout.

\subsection{Context and Training Objective}
\label{sec:training_details}

\paragraph{Context windows.}
Action pretraining packs consecutive observation--response pairs into
20-round windows. Instruction tuning uses up to 16 rounds and, when advancing
a full window, retains the latest eight complete rounds before adding new
ones. The active instruction, retained local reasoning, and task memory
accompany this history. State availability follows the effective times
established in Section~\ref{sec:data_examples}.

In Figure~\ref{fig:training_stages}, each column pairs an observation with
its response, and the shared eight rounds align across the two windows.
Selected responses in those rounds are supervised in the first window
and supply input history in the next.

\paragraph{Supervision scope.}
Loss applies to selected assistant responses in the new rounds. System
and user messages, retained responses, task memory, and input reasoning
are masked. For a Brain request, the proposed next instruction and its
initial local reasoning are targets alongside the action.

The assistant prefix ends at \code{<think>}. To continue the current
plan, the Policy learns to emit \code{</think>} as its first target token
and then predict the action. For a reasoning update, supervision begins
with the first reasoning token and includes the closing tag and action.
In the recovery examples for consecutive empty actions, the expert continuation supplies the
targets. Appendix~\ref{app:training_samples} shows the message sequences.

\paragraph{Response objective.}
Both stages optimize autoregressive response prediction. For selected
target-token indices \(\mathcal{T}\), the objective is weighted cross-entropy:
\begin{equation}
    \mathcal{L}_{\mathrm{CE}} =
    -\frac{1}{Z}\sum_{j\in\mathcal{T}}
        w_j\log p_\theta(y_j\mid y_{<j},X),
\end{equation}
where \(X\) is the visual and textual context, \(w_j\) is the token weight,
and \(Z\) counts the supervised target tokens across the global batch before
weighting.

Gameplay contains long runs of unchanged controls. In instruction tuning,
whole-action weighting reduces the contribution of repeated recorded actions:
\begin{equation}
    w^{\mathrm{action}}_t=\gamma^{k_t-1},
    \qquad \gamma=0.5,
\end{equation}
where \(k_t\) counts consecutive equal canonical ground-truth actions along
the continuous source sequence, including actions preceding the visible
window. Successive repeats receive action weights of 0.5, 0.25, and so on
after the first action at unit weight. Reasoning and Brain-payload tokens
keep unit weight. Appendix~\ref{app:training_details}
provides the two-stage training configuration.

Training and inference use matching assistant prefixes so that the Policy
learns the same continuation and update decisions it makes during deployment.

%% file: sections/06_realtime.tex
\section{Inference Infrastructure}
\label{sec:realtime}

\begin{figure}[!htb]
    \centering
    \input{figures/history_context_evolution.tex}
    \caption{Dual-route inference and scheduled control.
    (a) The retained prefix \(p_{17}\) contains context \(C\), complete rounds
    9--15, and \(u_{16}\). Completed response \(a_{16}\) and new observation
    \(u_{17}\) are appended before round-17 generation.
    (b) Two TP4 routes on one eight-GPU node alternate generation and prefix
    preparation while the preceding action executes. Solid arrows reuse a
    prepared prefix; dashed gray arrows carry the completed response.
    An early controller state waits for its assigned start (amber).
    The 200\,ms grid shows the control schedule; computation widths are
    schematic and the capture-to-submission offset is omitted.}
    \label{fig:history_policy}
    \label{fig:dual_routes}
\end{figure}
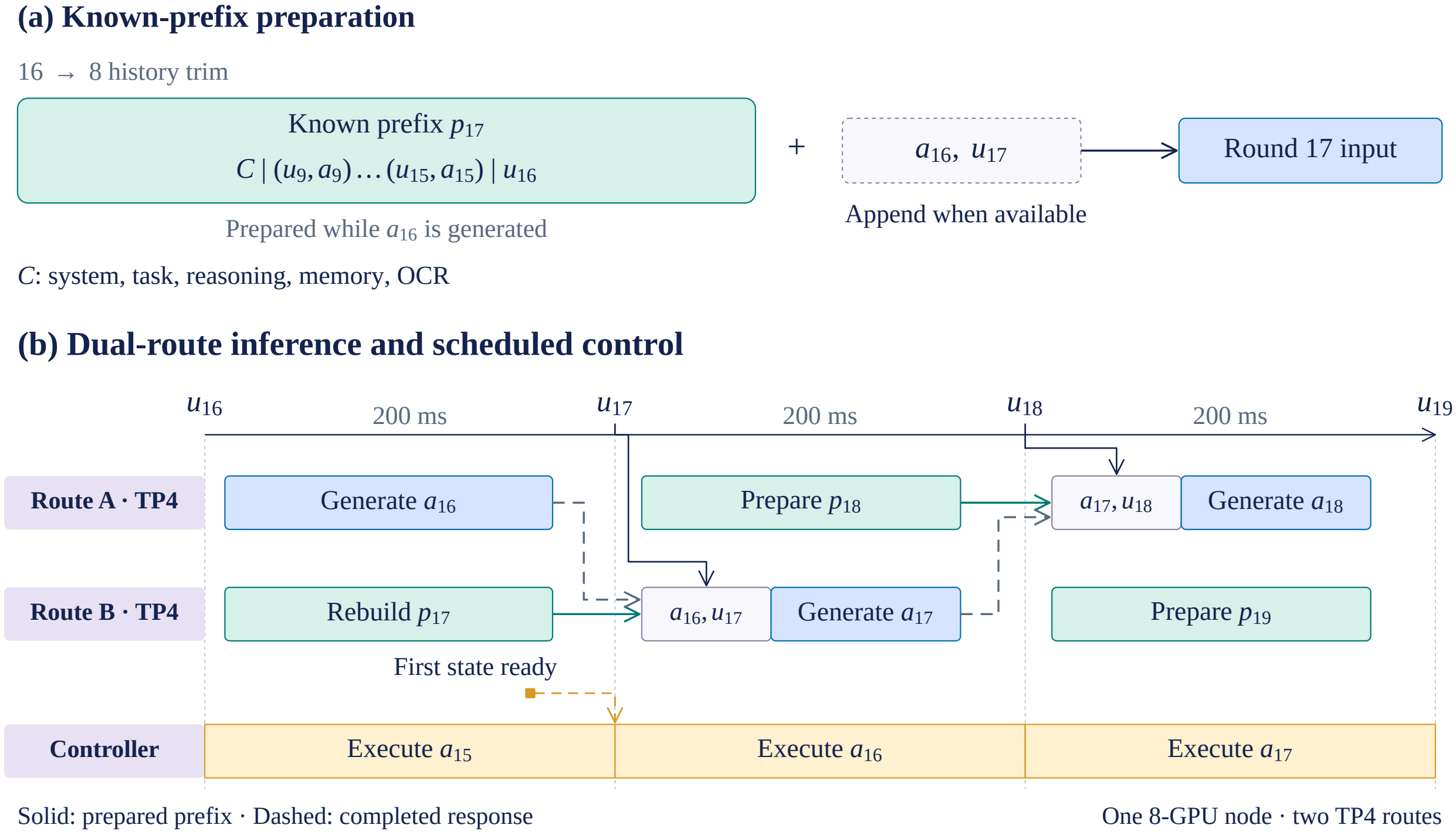

During play, we schedule Policy decisions at 5\,Hz to supply controls
for successive 200\,ms intervals. Once the history window is full, the
Policy trims its history every eight decisions. On a single inference
route, rebuilding the retained context delays the response on those
rounds. We therefore
alternate between the two routes shown in Figure~\ref{fig:history_policy}(b).
One prepares the context for the next decision while the other generates
the current response and the preceding action executes.

\subsection{Dual-Route Inference}
\label{sec:history_policy}
\label{sec:dual_routes}
Training and serving share a \((W,K)=(16,8)\) history window. After 16
completed rounds, the latest eight are retained. Each round pairs the full
frame and minimap shown in Figure~\ref{fig:data_example} with the Policy
response. Task context and delivered OCR persist across visual-history trims.

Trimming alters the context underlying attention KV and recurrent state.
Lumine combines StreamingLLM cache reuse with reasoning-triggered context resets~\citep{tan2025lumine}, while showing the limitation that KV cache is not exactly the same as that from re-calculated, resulting 8\% wrong actions\footnote{Measured individually.}. To solve the problem, we reconstruct the selected context
exactly and overlap this work with generation on a second route.

While one route generates \(a_t\), the other prepares \(p_{t+1}\): retained
history through \(u_t\), excluding unfinished \(a_t\) and future
\(u_{t+1}\). The next request extends this prefix with the completed
response and new observation, and the routes exchange roles. Reuse
requires matching prefix tokens and ordered media; a mismatch invokes
exact prefill. New-image upload overlaps unfinished prefix preparation;
reuse waits for both to complete.

Each route uses four-way tensor parallelism (TP4) and maintains its own
attention and recurrent caches. Both routes fit on one eight-H200 node.

\label{sec:history_latency}
\label{sec:window_results}
A \((W,K)\) window trims every \(W-K\) decisions. Retaining more history
at fixed \(W\) increases prefill size and reconstruction frequency;
16/8 provides eight decisions between trims. We choose this setting to
preserve recent context, limit reconstruction frequency, and leave
latency margin within the 200\,ms control interval.
See Appendix~\ref{app:history_retest} for detailed comparisons and
experimental settings.

\paragraph{Replay latency.}
In separate replays on two H200 nodes, dual routes reduce 95th-percentile request latency
from 230.0--230.9\,ms to 162.5--171.5\,ms. Table~\ref{tab:infra_dual_route_replay}
compares one TP4 route with two alternating TP4 routes running Qwen3.5-9B
in BF16 on the same 160-frame sequence with autonomous responses. The
largest gain comes on the rounds that rebuild history, where mean
latency falls from 218.1--221.0\,ms to 133.2--136.1\,ms.
Other rounds take slightly longer, leaving the overall means similar.
Timing runs through response completion.

\input{tables/infra_dual_route_replay}

\paragraph{Handoff after a task-state update.}
When local reasoning or task memory changes, the active pair continues to
supply controls while a standby pair prepares a fresh session. The harness
described in Section~\ref{sec:harness} transfers control once the standby
pair's first action and continuation prefix are ready. Placing the standby
pair on a second node gives it independent GPU resources.

In the controlled state-refresh replay described in Appendix~\ref{app:infra_handoff},
moving background preparation to a second node reduced mean handoff readiness
from 364.1 to 265.2\,ms. The p95 of the longest gap between available action
states per trial fell from 269.5 to 215.4\,ms. Immediate single-node reset
was ready in 244.4\,ms on average and had a larger p95 gap of 378.9\,ms.

\FloatBarrier
\subsection{Policy Inference Optimization}
\label{sec:engine_optimizations}
\label{sec:infra_benchmarks}
A Policy request combines visual encoding, multimodal prefill, and
autoregressive decoding. A prepared prefix lets the model reuse history
states and process the newly appended suffix before generating the next
response. Our serving path reduces overhead in model execution, image
processing, and request scheduling. We evaluate these components with
Qwen3.5-9B in BF16 on H200 GPUs, using fixed 20-token outputs. Every request contains 15 completed history
rounds and the current observation, with a full frame and minimap per round.
We use 16 scenes, repeat each five times on each of two nodes, and hold
these inputs fixed across all configurations. The visual cache holds
32,768 embeddings, covering the full 17,152-embedding window. Fixed output
length isolates serving cost; deployed responses stop when their action is complete.

\begin{figure}[!t]
    \centering
    \includegraphics[width=\linewidth]{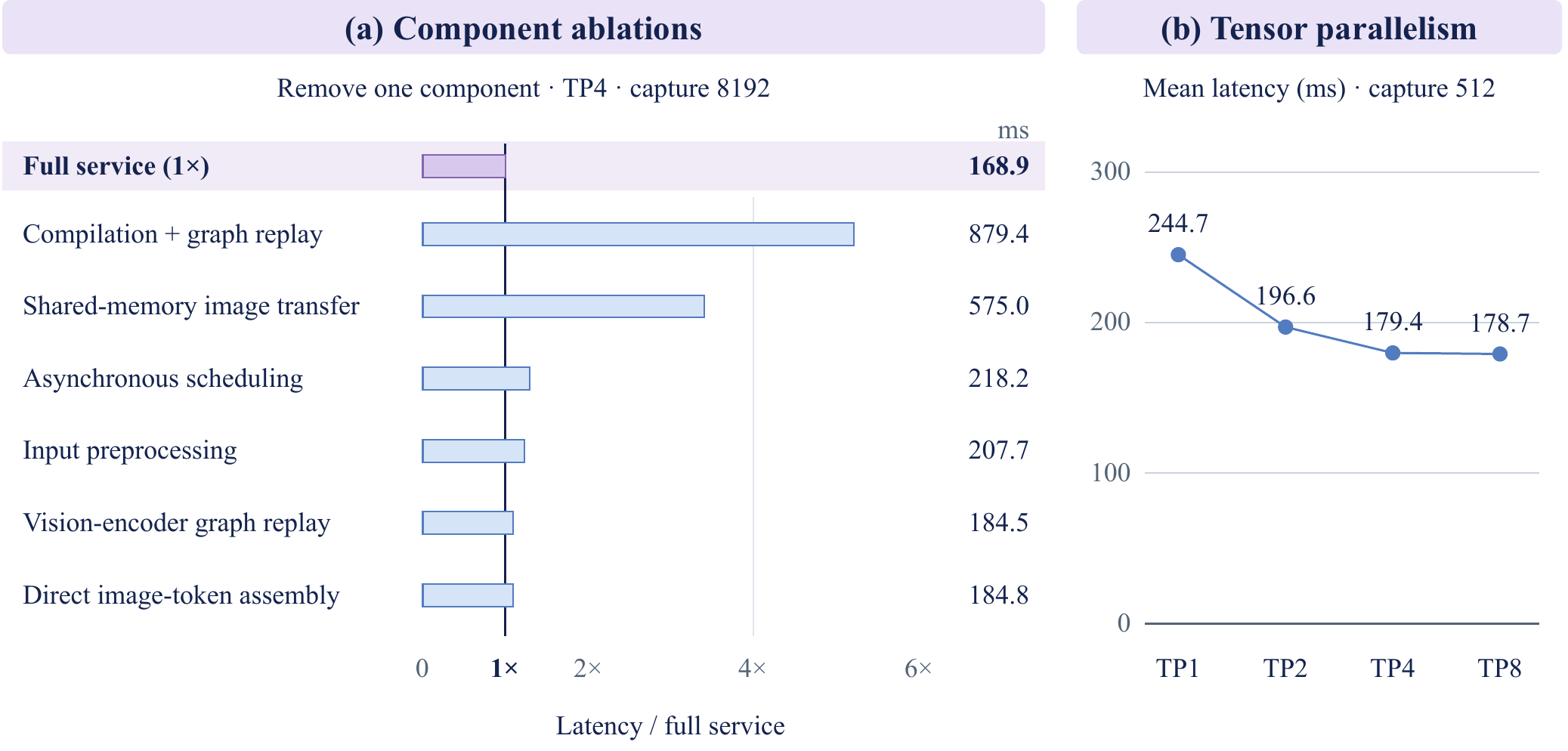}
    \caption{Policy inference latency for Qwen3.5-9B on H200 GPUs in BF16, with 15 history rounds,
    one current observation, and 20 output tokens. (a) Removing compilation and graph
    replay raises mean TP4 latency from 168.9 to 879.4\,ms. Removing shared-memory image
    transfer raises it to 575.0\,ms. Bar lengths show the mean latency of each
    configuration relative to the full service. (b) With a shared 512-token capture limit,
    TP4 and TP8 average 179.4 and 178.7\,ms. Panel (a) uses an 8192-token limit. Both
    panels use the same 16 scenes, five repetitions per scene on each of two nodes, and a
    32,768-embedding visual cache. Timing excludes upload and prior prefix preparation.
    Table~\ref{tab:infra_policy_efficiency} gives results for each node.}
    \label{fig:infra_policy_efficiency}
\end{figure}

\paragraph{Model execution.}
Short action responses repeatedly invoke the decoder with a small batch,
so host dispatch and repeated execution setup contribute to request
latency. Our vLLM-based service combines compiled execution with CUDA
Graph replay to reduce this overhead. Graph replay also covers the vision
encoder, which processes the full frame and minimap for each observation.
The complete TP4 service in Figure~\ref{fig:infra_policy_efficiency}(a)
averages 168.9\,ms per request. Replacing compiled
and graph execution with eager execution raises this to 879.4\,ms;
using the standard vision-encoder execution path raises it to 184.5\,ms.

\paragraph{Multimodal input processing.}
Shared-memory image tensors reduce repeated serialization and transfer
between the processor and inference engine while allowing prepared inputs
to be reused. Replacing this path with process-local LRU caching raises
mean latency to 575.0\,ms.

The input processor inserts image tokens directly into the prompt and uses
a fast lossless PNG encoder for the same-frame minimap. The decoded RGB
pixels are preserved. Figure~\ref{fig:infra_policy_efficiency}(a) gives
the latency of the individual preprocessing controls.

\paragraph{Request scheduling.}
Asynchronous scheduling overlaps host-side request handling with GPU
execution. Session-scoped caching also lets each route reuse its prepared
attention and recurrent states. In the history-length sweep shown in
Figure~\ref{fig:infra_history_length} and detailed in
Appendix~\ref{app:history_length}, a request with 15 history rounds takes
353.8\,ms when the text state is rebuilt during the request and 168.9\,ms
when the prefix is prepared beforehand. The replay in
Section~\ref{sec:dual_routes} measures how this preparation affects latency
across successive decisions.

\paragraph{Tensor parallelism and overall latency.}
Figure~\ref{fig:infra_policy_efficiency}(b) shows the optimized serving path
across tensor-parallel degrees. We fix the capture limit at 512 tokens
for all TP degrees because the 8192-token configuration exceeded GPU memory
during TP1 startup profiling. The scaling comparison uses the same
full-window inputs and repetitions as the component controls. At TP1,
the optimized service reduces mean latency from 620.9\,ms to 244.7\,ms
(2.54\(\times\) by the ratio of means). With the same capture limit,
TP2, TP4, and TP8 average 196.6, 179.4, and 178.7\,ms, respectively.
We use TP4 so that one eight-GPU node can host two independent routes.

Expanding CUDA Graph coverage from a maximum execution size of 512 to
8192 tokens reduces TP4 latency from 179.4\,ms to 168.9\,ms on the same
inputs, a 5.9\% reduction. The expanded coverage is used in the deployed
TP4 configuration. Detailed protocols and window comparisons appear in
Appendix~\ref{app:infra_details}.

\paragraph{Deployment configuration.}
Our default Policy deployment runs Qwen3.5-9B in BF16 on one eight-H200
node, with two independent TP4 routes using the full optimized serving
path and expanded CUDA Graph coverage. The routes alternate generation
and exact prefix preparation with a 16/8 history window. Visual-feature
cache capacity covers the full window, and observations and action delivery
follow the 200\,ms schedule described below. An optional second eight-H200
node hosts another pair of TP4 Policy routes with the same model and
inference configuration. These routes prepare an updated session while the
active node continues control; handoff follows the readiness checks in
Section~\ref{sec:harness}.

\FloatBarrier
\subsection{Real-Time Scheduling}
\label{sec:policy_scheduling}
Observations follow a 200\,ms capture grid, and each request receives an
absolute action-start time and a 200\,ms execution window. The harness
streams controller states during decoding, as described in
Section~\ref{sec:harness_policy_execution}. The scheduler places each
state within the assigned window. Early
states wait until the scheduled start; late arrivals begin execution on
arrival and have their tails cut off at the original window end. Expired
states are discarded. The preceding action
can therefore execute while the next observation is processed and the
next response is generated, as shown in Figure~\ref{fig:history_policy}(b).

Request scheduling respects the harness's response-completion dependency:
each dialogue admits one generation at a time and skips missed submission
slots to avoid accumulating stale requests. Within an admitted request,
new-image upload and background prefix preparation overlap; inference
begins once both are ready. The two routes alternate generation and
preparation while action execution follows its own scheduled window.
Appendix~\ref{app:control_scheduling} describes controller-state retention
across handoffs and watchdog behavior.

%% file: figures/history_context_evolution.tex
\includegraphics[width=\linewidth]{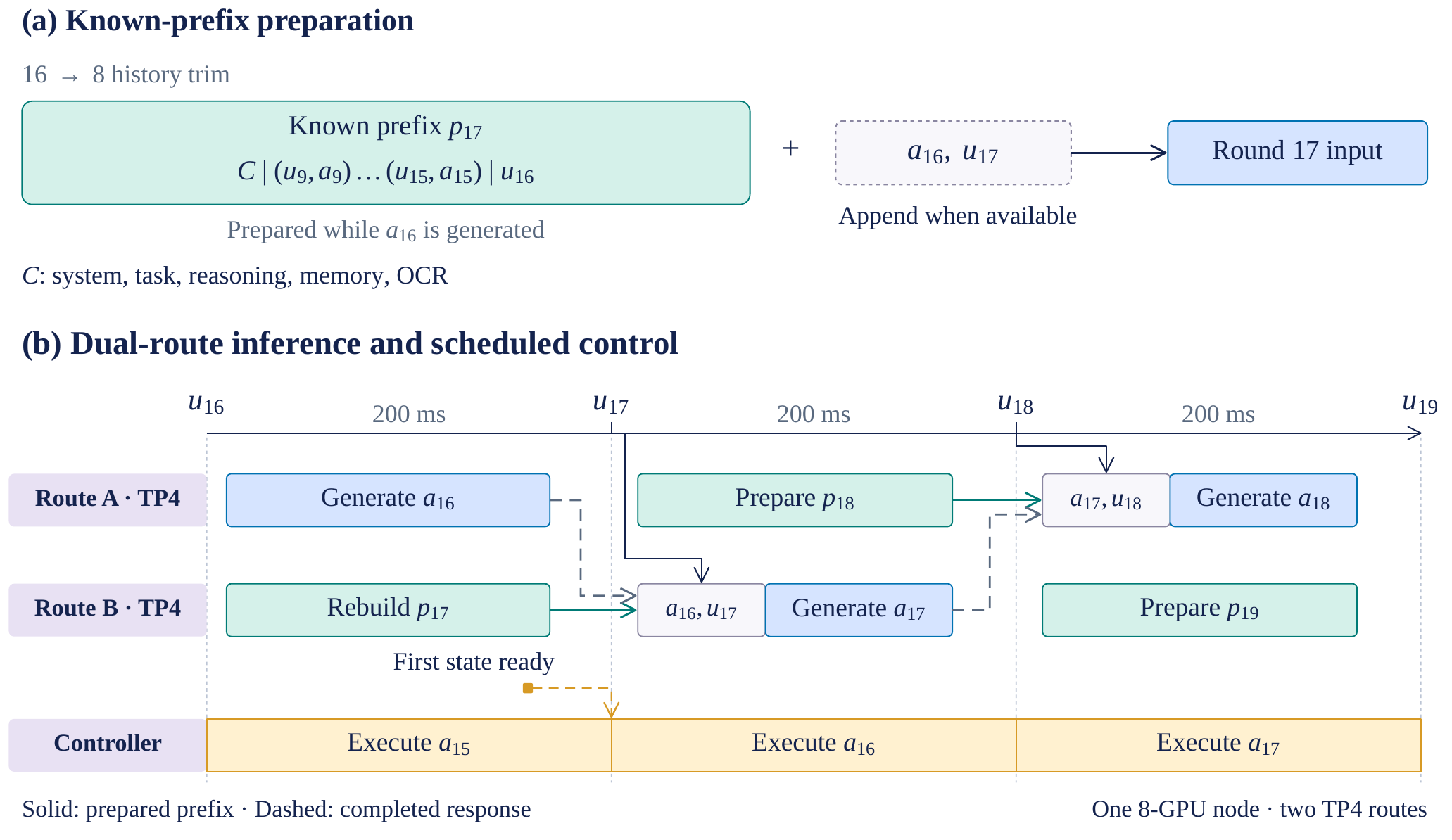}

%% file: tables/infra_dual_route_replay.tex
\begin{table}[!htbp]
\centering
\small
\setlength{\tabcolsep}{8pt}
\renewcommand{\arraystretch}{1.12}
\begin{tabular}{lrrr}
\toprule
Policy routes & Non-trim mean & Trim mean & Overall p95 \\
 & (ms) & (ms) & (ms) \\
\midrule
Serial TP4 (4 GPUs) & 149.9--150.6 & 218.1--221.0 & 230.0--230.9 \\
Dual TP4 (8 GPUs) & 157.3--161.8 & 133.2--136.1 & 162.5--171.5 \\
\bottomrule
\end{tabular}
\caption{History reconstruction and request latency at 16/8 (maximum 16 rounds, retain eight).
Both conditions use BF16, a graph-capture limit of 8192 tokens, and a visual-cache
budget of 32,768 embeddings per route.
Ranges span two independently tested nodes, each with three 160-frame replays.
After excluding the first 16 decisions of each replay, each configuration has
432 requests per node: 378 non-trim and 54 trim requests. Dual routes move
reconstruction off the trim request path. Timing includes image preparation, upload,
foreground prefix work, and autonomous response decoding; native action execution is excluded.}
\label{tab:infra_dual_route_replay}
\end{table}

%% file: sections/06_harness.tex
\FloatBarrier
\section{Agent Harness}
\label{sec:harness}

\begin{figure}[!t]
    \centering
    \includegraphics[width=\linewidth]{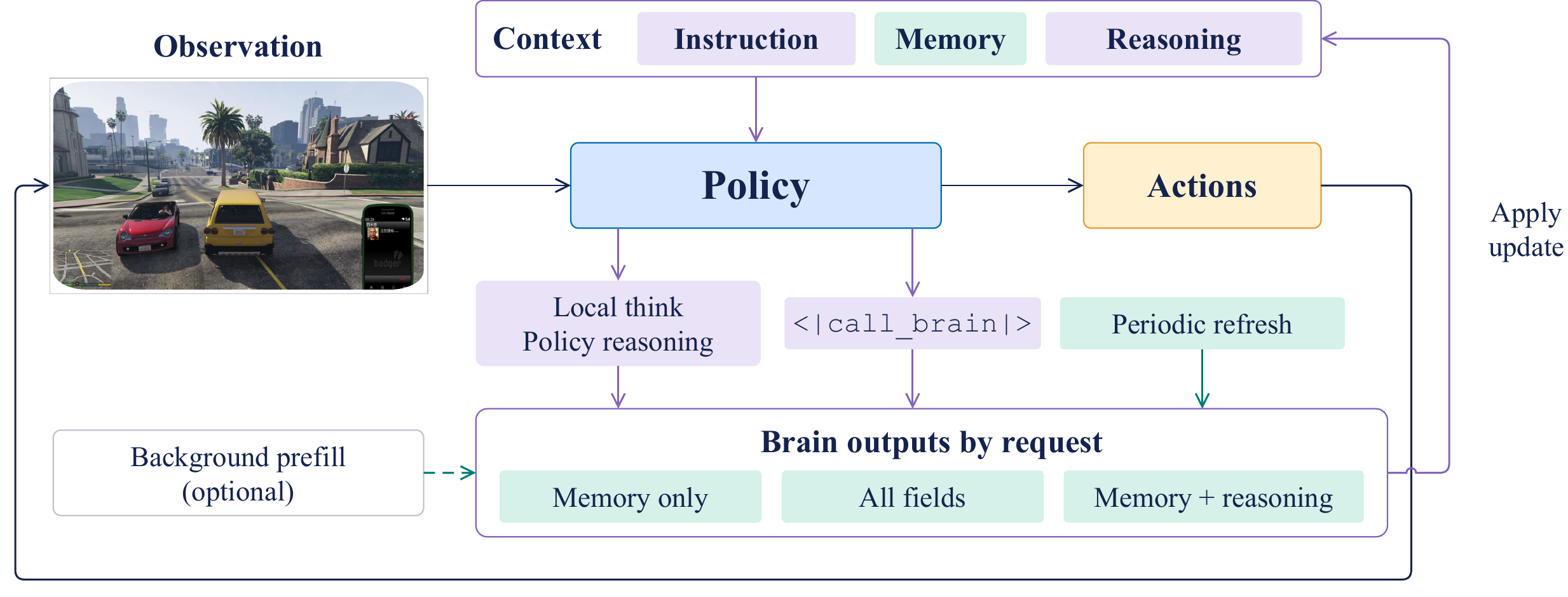}
    \caption{Online control and context updates in the agent harness. The Policy issues actions
    from the current observation and task context. When the Policy revises its local
    reasoning, the Brain updates memory. A full Brain call pauses play while the Brain
    reviews the instruction, memory, and reasoning. During a periodic refresh, the Brain
    updates memory and reasoning under the current instruction. Local and periodic updates
    can be prepared while the Policy continues to act. A self-hosted Brain can optionally
    prefill the shared evidence before a request.}
    \label{fig:harness}
\end{figure}

As shown in Figure~\ref{fig:harness}, the harness keeps the Policy in
control while a local update is prepared. When the Policy requests a stage
review, the harness pauses the game and resumes play with the reviewed
task state. It connects these decisions to the game by assembling model
inputs, recording which controls execute, and activating accepted updates.

Each Brain request includes the user-level task, previous task state, and
a chronological selection of screenshots, OCR, and action-execution evidence.
The input is frozen at a defined evidence cutoff. A Policy-triggered request
also includes its source observation, judgment, and any proposed next stage.
Brain's evidence window is maintained separately from Policy's short dialogue
window, retaining boundary frames and prioritizing changes in task text or
task state within the input budget.

\subsection{Policy Inference}
\label{sec:harness_policy_execution}

At each decision, Policy receives the active context
\(c_t=(g_t,m_t,r_t)\) from Section~\ref{sec:agent}: stage instruction,
task memory, and local reasoning. Its input comprises the current
frame and same-frame minimap, delivered OCR, and recent observation--response
pairs. OCR runs asynchronously: each request uses results from the preceding
submitted observation only if they are ready at submission. Late results
are omitted from Policy input, and recognition failures do not clear
previously delivered text. The current profile retains eight rounds when the
16-round window advances. Task memory persists beyond this visual window;
history eviction does not itself update the task state.

A fresh compact-Policy session uses the training-compatible unknown-goal
initialization with empty task memory and local reasoning. Subsequent
observations establish the current stage. Later context rebuilds preserve
the accepted task state.

Policy generates its reasoning decision and action in one response. The
harness incrementally forwards complete controller states to the executor
while decoding continues. Once the full response passes format validation,
it enters the Policy dialogue and can condition the next request without
waiting for physical execution to finish. An empty think continues control
with the retained local reasoning and does not itself invoke Brain. Nonempty
updates and Brain calls initiate the processes below. The response format
is specified in Section~\ref{sec:data_examples}; execution timing follows
Section~\ref{sec:policy_scheduling}.

Generated dialogue and actual execution are recorded separately. A complete,
valid response remains in Policy history even if its action executes only
partially or expires. Brain's trajectory evidence uses execution receipts
to distinguish completed actions, interrupted segments, and actions still
running at capture. It assesses outcomes from visual evidence and retained
facts; sending a control input alone does not establish task progress.

\subsection{Local Updates}

Local updates revise the current judgment and task memory while keeping
the stage instruction fixed. They have two sources, which differ in who
generates the next local reasoning.

\paragraph{Policy-triggered updates.}
A complete, valid response containing nonempty think without
\code{<|call_brain|>} supplies a new local judgment \(r_\pi\).
The harness sends this judgment and its supporting trajectory to Brain,
which returns updated task memory \(m_B\). The pending next context is
\(c^{\mathrm{next}}=(g_t,m_B,r_\pi)\): reasoning comes from Policy,
and task memory comes from Brain.

\paragraph{Periodic refresh.}
Even when Policy continues producing empty think, the harness can request
a refresh after a configured interval without a successful task memory
update. In this fallback request, Brain generates both task memory \(m_B\)
and local reasoning \(r_B\) from the recent trajectory. The pending context
is \(c^{\mathrm{next}}=(g_t,m_B,r_B)\), preserving the current stage
instruction. The rollout profile specifies the refresh interval; the
request is scheduled by the harness and leaves Policy's recorded response
unchanged. Profiles that enable a full-state fallback declare that scope
separately, since it also permits reconsidering the stage instruction.

Both local-update paths can run alongside Policy inference. In the handoff
configuration, the active Policy service continues control with \(c_t\)
while Brain computes its response. Once that response is validated, a
standby service with the same Policy weights prepares an independent session
for \(c^{\mathrm{next}}\). It starts with the revised state and an empty
visual dialogue, then generates a candidate from a fresh observation. The
active context remains unchanged until the standby is ready.

The harness transfers control after the standby produces a complete, legal
action and confirms the corresponding context and continuation-prefix
readiness within the action's execution window. Task state and Policy
session become active together, preserving held controls when requested
by the new action. If standby preparation fails, the active service retains control under
the existing protection rules, and the harness discards stale updates.

\subsection{Stage Replanning}

A complete, valid Policy response containing \code{<|call_brain|>}
requests a full Brain update. In the post-call format, Policy also proposes
the next stage instruction and initial local reasoning. The harness stops
new Policy submissions, settles or cancels outstanding actions, releases
inputs, and confirms that the game is paused. Pending local updates from
the previous context are invalidated.

Brain reviews the frozen trajectory, current task state, and Policy's
proposal to reconsider the active requirement. The proposal is a model
judgment to be checked against the screenshots, OCR, and execution evidence.
Brain returns the stage instruction, task memory, and initial local
reasoning, yielding \(c^{\mathrm{next}}=(g_B,m_B,r_B)\). This reassessment
may confirm the existing instruction when the evidence does not support
a stage transition.

After validating Brain's current response, the harness commits the three
fields, rebuilds the Policy session, and resumes from a new observation.
Policy's original proposal remains recorded.

\paragraph{Background prefill.}
When a self-hosted Brain supports background prefill and prefix reuse, the
harness can prepare stable instructions and known trajectory evidence while
Policy acts. A formal request appends the latest observations, current task
state, and update type to a valid completed prefix. Reusing the cached prefix
reduces the remaining prefill work for Policy-triggered memory updates,
periodic memory-and-reasoning refreshes, and full \code{<|call_brain|>}
requests. The requested outputs are generated after the trigger using the
updated evidence.

%% file: sections/10_hardware.tex
\section{Hardware}
\label{sec:hardware}

To deliver ZETA's gamepad actions to an Xbox console, a host computer sends
serial commands to an ESP32-P4 bridge, which converts them into controller
reports for the console. During autonomous operation, all gameplay commands
come from the computer; a genuine Xbox controller remains connected for
live authentication. The Policy supplies gamepad segments on a nominal
5\,Hz schedule, with each segment covering 200\,ms. To sustain the commanded state during execution, the
host can refresh it over the serial link at the 50\,Hz rate recommended
by the V1 protocol.

\subsection{USB Connection and Authentication}
\label{sec:controller_hardware}

The bridge has three connections: the genuine controller connects to its
USB Host port, the Xbox console connects to its USB Device port, and the
computer connects over UART or a USB-to-UART adapter. Controller
initialization precedes the device connection to Xbox. The board then
passes authentication traffic between the genuine controller and console,
preserving the controller as a live authentication source throughout use.
Serial actions are accepted once the connection reaches \code{READY};
commands received before authentication are rejected.

\subsection{Controller Signal Translation}

The computer sends one text command per line. Commands specify buttons, the
directional pad, both sticks, and both triggers. For coordinated input,
\code{set} validates an entire command before updating its fields together
in firmware state. For example, \code{set ls=0,-70 rt=100 ms=500} combines
a left-stick position and a fully pressed right trigger for 500\,ms.

The firmware parses each command, updates the requested controller state,
and constructs the outgoing Xbox report. Stick commands use integer
percentages from $-100$ to $100$, mapped to signed 16-bit Xbox
axis values with the corresponding Y-axis convention. Trigger percentages
from 0 to 100 map to the console's 0--1023 range. A multi-field update is
atomic in firmware state; USB report transmission follows its own schedule.

\subsection{Timed Execution and Feedback}

Commands assign finite control durations of 20--2000\,ms. A new command
replaces the affected state and expiry time, so the computer schedules the
action sequence and refreshes sustained controls. The board holds each state between refreshes. Expired serial commands are
cleared; \code{neutral} sets supported controls to neutral values for
the specified duration.

Session identifiers and sequence numbers reject stale commands and prevent
duplicate execution on retries. Disconnecting, suspending, or reauthenticating
clears serial actions without replay. Replies acknowledge accepted commands,
while events report serial state changes, expiry, and tap completion.
These signals let the computer track the
controller interface; visual feedback establishes the resulting in-game
behavior, since command acceptance alone does not confirm a game response.
\FloatBarrier

%% file: sections/07_experiments.tex
\section{Evaluation}
\label{sec:experiments}

We evaluate ZETA through real-game task performance in GTA V, training ablations,
and failure analysis. The performances in GTA VI are shown in our project home.

\subsection{Real-Game Evaluation}
We select 10 main-story missions in GTA V, divide them into 47 subtasks,
and conduct real-game evaluation. Each subtask starts independently
from its fixed game state.

Each episode specifies a fixed initial state, task objective, interaction
budget, and terminal success evidence. Success is determined by native
game events confirming progress to a predefined subsequent stage or
successful mission completion.
Failures and timeouts count as unsuccessful attempts.
The inference runtime and controller interface are described in
Sections~\ref{sec:realtime} and~\ref{sec:hardware}, respectively.

We use Kimi-K3 as the Brain for high-level reasoning and memory updates.
The \texttt{MOSS-ZETA-preview} Policy is deployed as two tensor-parallel
routes (TP4 each) on one eight-H200 node per game. The routes alternate
action generation and context prefill, as described in
Section~\ref{sec:realtime}.

For \texttt{MOSS-ZETA-preview}, we report success rate at the subtask level. \texttt{MOSS-ZETA-preview} operates within the harness
and infra described above. Each subtask receives five attempts, with a
maximum of 20 minutes of game time per attempt. We also evaluate
\texttt{gpt-6.2-sol} as a direct visual gamepad controller with medium
reasoning effort. It receives the task objective, current screenshot, and
recent observation--action history, and predicts a one-second action
package containing joystick inputs and six sequential button-state
segments, each lasting approximately 166\,ms. These runs use a 16/8 history
window and a maximum of 1,200 action packages per attempt.

On this 47-subtask suite, \texttt{MOSS-ZETA-preview} achieves
66.0\% success rate and 80.9\% success rate for five times, while  \texttt{gpt-6.2-sol} as a direct visual gamepad
controller gets 36.17\% in the same setting and 74.47\% when pausing the game when the model is reasoning.

Beyond the selected 10 missions, we evaluate \texttt{MOSS-ZETA-preview}
on our complete main-story task suite, covering 69 missions and 394
independently initialized subtasks. With one attempt per subtask under the
same interaction budgets, the agent successfully completes 239 subtasks,
achieving a single-attempt success rate of 60.7\%.

\subsection{Ablation Experiments}
Starting from the same pretrained checkpoint, we construct training
variants using different data and loss settings. All ablation variants
use training data drawn only from the 10 selected main-story missions.
We compare training with and without the data augmentation described in
Section~\ref{sec:data_examples}, and with and without the repeated-action
loss decay defined in Section~\ref{sec:training_details}.
Table~\ref{tab:training_ablations} reports their single-attempt
success rates on the same 47-subtask suite. The augmented variant's rate
is averaged over its five attempts per subtask.

\begin{table}[htbp]
    \centering
    \small
    \begin{tabular}{lccr}
        \toprule
        Training variant & \shortstack{Data\\augmentation}
        & \shortstack{Repeated-action\\loss decay} & Success (\%) \\
        \midrule
        Baseline & Off & On & 42.55 \\
        Augmented & On & On & \textbf{54.89} \\
        No repeated-action loss decay & Off & Off & 4.26 \\
        \bottomrule
    \end{tabular}
    \caption{Data augmentation and repeated-action loss decay comparisons
    on the same 47-subtask suite. The reported metric is single-attempt
    success rate.}
    \label{tab:training_ablations}
\end{table}

\paragraph{Data augmentation.}
The augmented model obtains 54.89\% single-attempt success, compared
with 42.55\% for the baseline, an improvement of 12.34 percentage points.

\paragraph{Repeated-action loss decay.}
The variant without repeated-action loss decay completes 2 of the
47 subtasks, achieving 4.26\% success, compared with 42.55\%
(20 of 47) for the baseline that retains it. Together, these results show
that both data augmentation and the repeated-action weighting in the
training loss improve real-game task success.

\subsection{Failure Analysis}
We review archived observations and executed actions from the earlier
augmented model's evaluation. The cases highlight two limitations:
consistent control in highly dynamic tasks and recovery after local errors
or stalled progress.

\paragraph{Highly dynamic tasks.}
Pursuit tasks require continuous adaptation to a moving target. When the
target turns, passes obstacles, or temporarily leaves view, the previous
movement direction may quickly become outdated, requiring the agent to
locate the target again and adjust its controls. In a reviewed foot chase,
the target moved deeper into a bus depot while the character repeatedly
adjusted its position and camera near a building and a bus wreck, without
maintaining reliable pursuit. The attempt ended in failure. Local errors
can compound in such tasks: the target continues moving while the
character reorients or negotiates an obstacle, making reacquisition and
pursuit more difficult. Although individual movement and turning actions
can be executed, the agent does not yet consistently maintain continuous,
timely target tracking as the environment changes.

\paragraph{Recovery capability.}
After a local error, the agent may continue acting without consistently
restoring task progress. In a reviewed driving attempt, a damaged vehicle
repeatedly changed heading on the same elevated road segment and made
contact with roadside barriers or other vehicles. Subsequent adjustments
did not restore stable route following before timeout. In a clothing-store
attempt, the character moved among clothing racks and tried inputs but did
not open the purchase menu before exhausting the task budget. These cases
illustrate recovery difficulties in locomotion and scene interaction:
the former requires correcting the vehicle's position and heading after
collisions, while the latter requires re-establishing the conditions for
a successful interaction. Current corrective behavior does not always
end stalled progress effectively, allowing a local error to develop into
prolonged ineffective control.

%% file: sections/08_related_work.tex
\section{Related Work}
\label{sec:relatedwork}

\subsection{Game Agents}
\label{sec:related_game_control}

Games have supported advances in learned decision making, from visual
control in Atari to competitive play in Dota 2 and StarCraft
II~\citep{mnih2015dqn,berner2019dota,vinyals2019alphastar}.
Research in open-world environments broadens this agenda toward varied
goals and reusable behavior. Minecraft agents draw on gameplay video,
Internet-scale multimodal knowledge, and instruction-conditioned
policies~\citep{baker2022vpt,fan2022minedojo,lifshitz2023steve1}.
VPT supplies a behavioral prior by labeling video with an inverse dynamics
model, MineDojo connects an open-ended task suite to a multimodal knowledge
base, and STEVE-1 adapts a pretrained controller to text and visual goals.

Recent visual agents connect instructions to low-level inputs across
commercial games and simulated worlds~\citep{sima2024scaling,sima2025sima2,tan2025lumine,ouyang2026player1}.
Within this line, SIMA 2 incorporates foundation-model reasoning and
self-improvement; Lumine combines adaptive reasoning with real-time
control for hours-long missions. Player-1 investigates synthetic-environment
reinforcement learning as a way to improve control beyond demonstrations.
GTAGym, supplies executable GTA V tasks and outcome
verification for evaluation and reinforcement learning~\citep{chen2026gtagym}.
ZETA develops sustained GTA V main-story execution through synchronized
gamepad demonstrations, task-state supervision, and coordinated Policy
and Brain inference.

\subsection{Reasoning and Memory for Long-Horizon Tasks}
\label{sec:related_reasoning_memory}

Reasoning and retained experience help agents revise plans as interaction
unfolds. Language-agent and embodied-planning methods use observations,
success signals, or verbal feedback to guide subsequent
decisions~\citep{yao2022react,huang2022innermonologue,shinn2023reflexion}.
ReAct interleaves reasoning and actions, Inner Monologue feeds environment
feedback into language-based planning, and Reflexion stores verbal
reflections across attempts.

Open-world agents extend these ideas through task decomposition, reusable
skills, and persistent memory~\citep{zhu2023gitm,wang2023voyager,wang2023jarvis1,li2024optimus1,tan2024cradle}.
GITM uses text-based knowledge and memory for Minecraft planning, while
Voyager accumulates executable skills through execution feedback.
JARVIS-1 connects multimodal memory and planning to goal-conditioned
controllers; Optimus-1 combines structured knowledge with summarized
multimodal experience. Cradle organizes screenshot interpretation,
planning, reflection, and memory around computer interaction.
ZETA retains stage instruction, local reasoning, and task memory alongside
a learned gamepad controller. Sparse Policy updates and external Brain
revisions let plans and grounded facts persist across action segments
and visual-history windows, while the harness determines when revised
state becomes active.

\subsection{Real-Time Vision--Language--Action Control}
\label{sec:related_realtime_vla}

Sequence modeling and multimodal pretraining underpin generalist control
policies across tasks and embodiments~\citep{reed2022generalist,brohan2022rt1,brohan2023rt2,octo2024policy,kim2024openvla}.
RT-2 and OpenVLA adapt vision--language representations to action
prediction, while Octo studies transferable policy initialization across
robot configurations. Predicting action sequences is another shared
design choice: ACT, Diffusion Policy, and \(\pi_0\) model coordinated
action chunks using different generative
formulations~\citep{zhao2023act,chi2023diffusion,black2024pi0}.
FAST further studies compressed action tokenization for autoregressive
VLAs~\citep{pertsch2025fast}. These works connect representation learning,
action serialization, and temporally extended prediction.

Real-time execution additionally requires managing inference delay and
continued context processing. RTC generates a new action chunk while
the current one executes, fixing committed actions and inpainting the
remaining sequence~\citep{black2025realtime}. StreamingLLM studies efficient
bounded-cache language generation by retaining attention sinks and recent
tokens~\citep{xiao2023streamingllm}. For ZETA, shifted action supervision
accounts for observation-to-execution delay, while dual inference routes
overlap exact reconstruction of the selected multimodal context with
response generation. The resulting autoregressive gamepad responses are
delivered through an explicit action schedule, connecting retained task
state and on-demand reasoning to continuous control.

%% file: sections/09_limitations_conclusion.tex
\section{Conclusion}
\label{sec:conclusion}
ZETA connects visual controller prediction with local reasoning,
higher-level updates, and persistent task memory for GTA V interaction.
Its training pipeline makes observation/action timing, annotation evidence,
and supervised response boundaries explicit. Its runtime separates delivered
observations, generated responses, accepted state updates, and confirmed
controller execution. The resulting technical account connects the training and runtime design
with real-game evaluation. On 47 subtasks from 10 main-story missions,
\texttt{MOSS-ZETA-preview} achieves 66.0\% \texttt{pass@1} and
80.9\% \texttt{pass@5}. The training comparisons further
motivate data augmentation and repeated-action loss decay in the default
recipe.

From GTA V to GTA VI, as games become increasingly realistic, large models are still able to retain strong generalization capabilities. Taking this one step further, we aim to explore the transition from virtual worlds to the physical world. Can an intelligent agent that understands its surroundings and accomplishes predefined goals in a virtual environment transfer its capabilities to real-world robots, ultimately bringing AGI into the physical world?
This is just a small first step toward that vision.

%% file: appendices/protocols_and_reproducibility.tex
\section{Action and Timing Protocol}
\label{app:actions}

Each 200\,ms source action contains six stick states. We retain the most
frequent joint tuple \((l_x,l_y,r_x,r_y)\), resolving ties by its earliest
occurrence. Each selected coordinate \(x\), expressed on the integer scale
\([-100,100]\), is quantized as
\[
    q=\operatorname{sign}(x)
      \left\lfloor\frac{9|x|}{100}+\frac{1}{2}\right\rfloor.
\]
After the data preprocessing described in Section~\ref{sec:data_alignment},
the four codes are serialized with an explicit sign, using \code{+0} for zero.

Each 100\,ms button block is normalized from three consecutive source
chunks. All-empty chunks remain empty. If there is one distinct nonempty
state, that state is retained. Otherwise, the union is chosen only when that
complete button state occurs among the original chunks; remaining cases use
the most frequent nonempty state, breaking ties by its latest occurrence.
The selected state is repeated across the three historical serialization
slots, yielding two ordered blocks over 200\,ms. This button normalization
preserves the four compact stick coordinates and is applied to both target
and historical actions.

The six serialized button slots thus represent two complete button sets,
one for each 100\,ms interval; an empty set denotes no pressed buttons.
The four stick coordinates and all six button slots together specify the
complete 200\,ms action segment.

Temporal alignment determines which recorded segment is paired with an
observation. The base conversion labels the interval
\((t,t+200\,\mathrm{ms}]\). The current delay-aware training target keeps
the observation at \(t\) and uses the complete next-round action for
\((t+200\,\mathrm{ms},t+400\,\mathrm{ms}]\), matching the prediction offset
and action duration \(\delta=\Delta=200\,\mathrm{ms}\) in
Section~\ref{sec:zeta_task}. Both stick coordinates and all button states
shift together, including historical responses. This mapping continues
across packed-window boundaries within a source; the final observation is
dropped when no next-round action exists. Serialization, parsing, button
normalization, and temporal alignment are recorded together with the export.

\section{Annotation Construction and Evidence Scope}
\label{app:annotations}

Each annotation artifact records the source recording, selected frame,
available image/OCR/action evidence, prior state, generated response, and
validation or review status. Stage-content review, alignment, and local reasoning
generation are different operations. In the documented frame-alignment
contract, semantic stage fields are inherited programmatically while
the alignment model supplies frame positions and their evidence.

Offline decision-point selection may inspect a longer interval to choose an
informative point. At the selected node, local reasoning may reference an
explicitly marked auxiliary block of subsequent images, OCR, and recorded
actions, bounded by three seconds and fifteen frames, to clarify the current
judgment and local plan. Task memory follows a causal content constraint:
it records facts established by the selected node from the selected segment
and preceding state. Later events must not be incorporated as already
completed progress.

The auxiliary future-evidence block is used only for offline annotation and
is excluded from Policy observation inputs during training and online
inference. Annotation activation follows the effective times in
Section~\ref{sec:data_examples}.

Task memory construction checks continuity as well as additions. A fact may remain
useful after an instruction change, while a stale plan may no longer be
appropriate despite being recently generated. The annotation rules cover
evidence, continuing relevance, invalidation, and the distinction between
a fact and an intended action.

\section{Training Prompts and Sample Organization}
\label{app:training_samples}

This appendix presents the controller-only and instruction-tuning system
prompts, followed by message organization and response targets. Both training
stages use Chinese system prompts; the corresponding English versions are
shown below. Braced expressions denote dynamic state values.

\subsection{Action-Pretraining System Prompt}
\label{app:pretraining_prompt}

The English version of the controller-only prompt used for action-data
preparation is reproduced below. It establishes the agent's identity,
gameplay objective, visual input, and serialized controller response.

\begin{Verbatim}
You are a GUI game-control agent developed by OpenMoss. You control a fullscreen GTA game using a gamepad.

You are acting as an experienced GTA player. Your high-level objective is to make progress in and complete the main story missions. Prioritize mission objectives, navigation markers, dialogue prompts, minimap guidance, and on-screen instructions over free roaming.

You will receive complete monitor screenshots of the game. Focus especially on the minimap. A yellow dot, yellow route, or yellow marker usually indicates the current mission objective. Sometimes it is purple. Based on the screenshot, decide the next gamepad actions to perform.

You must output exactly one 200ms gamepad-control sequence within <a> and </a>. The sequence must contain exactly six action chunks, each representing approximately 33ms, separated by exactly five semicolons ';'.

At the beginning of the sequence, output the stick position for the entire 200ms window exactly once as four fixed-width signed coordinate tokens in this order: left-stick x, left-stick y, right-stick x, right-stick y. Each token must be an explicit '+' or '-' followed by one digit from 0 to 9. A larger digit means the stick is pushed farther. For x, negative means left and positive means right. For y, negative means down and positive means up. A centered axis must be written as +0; never output -0. Concatenate the four tokens directly with no separators, so the coordinate prefix is always exactly eight characters, for example: -9+0+2+9.

Immediately after the fourth coordinate token, output the action tokens for the first 33ms chunk. After each semicolon, output the action tokens for the next chunk.

Available gamepad action special tokens are:
LT, LB, L3, view, up, down, left, right, RT, RB, R3, menu, X, Y, B, A

Within a 33ms chunk, concatenate all action tokens directly with no spaces or other separators. If a chunk has no action, leave it empty. Do not output L(...), R(...), parentheses, newline characters, commas, or spaces anywhere inside <a> and </a>.

Example output format:
<a>-9+0+2+9LT;LTRT;LT;LT;RT;RT</a>
\end{Verbatim}

\subsection{Instruction-Tuning System Prompt}
\label{app:instruction_tuning_prompt}

The instruction-tuning prompt adds local reasoning, Brain requests,
externally supplied task memory, OCR delivery, and retained history to the
controller instructions. The English translation below preserves control
tokens, JSON field names, and action examples. The three trailing state
values are replaced with placeholders.

\begin{Verbatim}
You are a GUI game-control agent developed by OpenMoss. You control a fullscreen GTA game using a gamepad.

You are acting as an experienced GTA player. Your high-level objective is to make progress in and complete the main story missions. Prioritize mission objectives, navigation markers, dialogue prompts, minimap guidance, and on-screen instructions over free roaming.

You will receive complete monitor screenshots of the game. Focus especially on the minimap. A yellow dot, yellow route, or yellow marker usually indicates the current mission objective. Sometimes it is purple. Based on the screenshot, decide the next gamepad actions to perform.

The action must be a short gamepad-control sequence inside <a> and </a>. It begins 200ms after the current screenshot and covers the following 200ms. It must contain exactly six action chunks: the first three represent the first 100ms and share the same button state; the last three represent the second 100ms and share the same button state. Separate the six chunks with exactly five semicolons ';'.

At the beginning of the sequence, output the stick position for the entire control sequence exactly once as four fixed-width signed coordinate tokens, in this order: left-stick x, left-stick y, right-stick x, right-stick y. Each token must be an explicit '+' or '-' followed by one digit from 0 to 9. A larger digit means the stick is pushed farther. For x, negative means left and positive means right. For y, negative means down and positive means up. A centered axis must be written as +0; never output -0. Concatenate the four tokens directly with no separators, so the coordinate prefix is always exactly eight characters, for example: -9+0+2+9.

Immediately after the fourth coordinate token, output the action tokens for the first chunk. After each semicolon, output the action tokens for the next chunk.

Available gamepad action special tokens are:
LT, LB, L3, view, up, down, left, right, RT, RB, R3, menu, X, Y, B, A

Controls: LT aims and RT fires; in vehicles they usually brake/reverse and accelerate, respectively. Use A for running, confirmation, or interaction according to the HUD. B reloads or performs melee actions, or goes back in menus, depending on the situation. X jumps or climbs; Y enters or exits vehicles; LB opens the weapon wheel; RB provides cover or the handbrake. Use L3/R3 according to the HUD. up opens the phone. Hold down to open the character-selection wheel and use the right stick to select; after confirming the highlight, leave the buttons empty in the next action while retaining the selected stick direction to release and confirm. Follow the current HUD for interactions rather than guessing buttons. Retrying and exiting after failure are different operations.

Within a chunk, concatenate all action tokens directly with no spaces or other separators. If a chunk has no action, leave it empty. Do not output L(...), R(...), parentheses, newline characters, or commas. Do not insert spaces anywhere inside <a> and </a>.

Example action format:
<a>-9+0+2+9LT;LT;LT;RT;RT;RT</a>

On each round, output <think>judgment</think><a>action</a> continuously. If there is no new judgment, output <think></think>. Do not add leading or trailing newlines inside think, or separator whitespace between </think> and <a>. Ordinary local judgments use natural language and do not reproduce or output Memory.

When the stage objective needs updating, first write a brief pre-call judgment inside think, then output <|call_brain|>, immediately followed by a JSON object containing exactly two fields: instruction (the next stage objective) and small_think (the initial local judgment for that objective).

The objective and initial local judgment take effect from the next round; accompanying Memory is provided externally. Ordinary local judgments are updated normally, while external Memory may lag the latest local judgment by one update point. Use the current screenshot to assess the situation; do not treat an older summary as certain evidence about the current screen. Retain the existing judgment when there is no local update. Do not copy control markers or pre-call reasons into the retained model reasoning. Major objective updates, initialization, and external fallback updates resynchronize state.

Memory is supplied only as system input and is not an output target of this model. The result of the current update becomes visible only from the next round.

OCR: Each record inside <screen_ocr> begins with S or D and is read in delivery order. In an S snapshot, omitted fields are empty. A D increment replaces only the listed fields; an empty value clears a field. Each round supplies the OCR block before the images. An empty block, <screen_ocr></screen_ocr>, supplies no changes; retain the existing OCR state. The image is the current observation, while the OCR in the same user message has a fixed one-round delay and comes from the preceding observation. It does not represent instantaneous text from the current image. At the first observation, when no preceding OCR is available, supply an empty S snapshot. After a context reset, establish the S snapshot from the already delivered OCR of the preceding observation. Retain historical OCR in delivery order. Similar recognition results may be merged; use the screenshot when assessing the current state.

Whenever image and action history reaches 16 rounds, retain the latest 8 rounds and continue accumulating. From evicted rounds, retain only changed OCR, aggregated in delivery order into a separate history message. Rolling eviction does not reset OCR state. After a state update, clear image, action, and OCR history from the next round and re-establish the OCR snapshot.

P = upper-left prompts; B = bottom task objectives or dialogue subtitles; C = central task outcomes; K = lower-right control instructions.
Separate multiple lines within one field with |. Escape literal | and backslash characters with a backslash.

Current effective objective: {instruction}

Memory: {memory}

Model reasoning: {model_thought}
\end{Verbatim}

\subsection{Message Organization}

The layouts below show message order. Media placeholders represent
processor-expanded visual inputs; English state-field labels correspond to
the translated prompt above.

Action pretraining places the controller prompt from
Section~\ref{app:pretraining_prompt} in the system message, followed by
alternating observations and recorded action responses. It does not add
the annotations for stage instructions, local reasoning, or task memory used
in instruction tuning. The following layout shows how the prompt and
observations are assembled:
\begin{verbatim}
system:    {action-pretraining system prompt}
user:      {observation_1}
assistant: {recorded action_1}
...
user:      {observation_n}
assistant: {recorded action_n}
\end{verbatim}
Consecutive pairs are packed into the pretraining windows described in
Section~\ref{sec:training_details}; later observations are causally hidden
when predicting earlier responses.

The instruction-tuning system prompt in
Section~\ref{app:instruction_tuning_prompt} ends with three dynamic fields:
the currently effective instruction, task memory, and retained local
reasoning. These fields describe the state available before the selected
response. A user message supplies delivered OCR followed by the current
full frame and same-frame minimap. Earlier user--assistant pairs precede
the selected response in temporal order:
\begin{verbatim}
system:    {fixed instruction-tuning prompt}
           Current effective objective: {instruction}
           Memory: {memory}
           Model reasoning: {model_thought}
user:      {delivered OCR and historical visual observation}
assistant: {historical response}
...
user:      {delivered OCR and current visual observation}
assistant: <think>{selected response continuation}
\end{verbatim}
Here the final \code{<think>} is the assistant generation prefix. The
selected response continuation begins with either \code{</think>} or new
reasoning. Earlier assistant responses and the dynamic system fields have
no target loss in this window.

\subsection{Input-Only Annotations and Response Targets}

For the annotation points defined in Section~\ref{sec:data_reasoning},
local reasoning at non-fallback decision points is supervised as a Policy
response. At fallback update points, the generated local reasoning serves
only as subsequent input context and receives no output supervision.
Recorded actions remain supervised in both cases.

Task memory is always supplied as input and is never a Policy output target.
It summarizes trajectory context that is not fully available in the Policy's
bounded visual history. Local reasoning from either type of point may later
be retained as input without receiving loss again in that role.

For the driving example, the input instruction asks the Policy to return
the recovered car to the dealership. Memory records the completed retrieval
and unfinished return trip; retained reasoning recommends following the
navigation route after the junction. A selected continuation is:
\begin{verbatim}
prefix: <think>
target: </think><a>-9-3+0+0LT;LT;LT;LT;LT;LT</a>
\end{verbatim}
The closing think token and action are supervised. The supplied memory and
retained reasoning are not. At a non-fallback local-update point, the response
begins with new reasoning, followed by the closing tag and action; a Brain
response additionally supervises its next-state payload.

When the window advances, selected responses may become retained history
and their loss is disabled in the new window. An externally refreshed input
state is visible only from its effective round, not retroactively to an
earlier target. For recovery from consecutive empty actions, five repeated observations paired with both
sticks centered and all buttons released form input-only history; the
unmodified expert continuation supplies the targets. Short initial windows and
state-reset segments need not contain eight historical rounds.

\FloatBarrier
\section{Additional Supervised-Training Details}
\label{app:training_details}

The Policy uses Qwen3.5-9B, comprising a visual encoder, a multimodal
projection, and an autoregressive language model. Action pretraining
initializes from pretrained multimodal weights. Both stages update the
visual encoder, projection, and language model with full-parameter training.
Instruction tuning initializes from the action-pretraining checkpoint at
step 3,500, with a fresh optimizer and learning-rate schedule.
Table~\ref{tab:recipe} summarizes the configuration of both stages. Both
use BF16 model parameters; instruction tuning uses FP32 main gradient
buffers and Adam moments.

\begin{table}[!htbp]
    \centering
    \small
    \begin{tabular}{>{\raggedright\arraybackslash}p{0.30\linewidth}
                    >{\raggedright\arraybackslash}p{0.29\linewidth}
                    >{\raggedright\arraybackslash}p{0.29\linewidth}}
        \toprule
        Setting & Action pretraining & Instruction tuning \\
        \midrule
        Initialization & Pretrained Qwen3.5-9B & Action-pretraining step 3,500 \\
        Adaptation & Full-parameter training & Full-parameter training \\
        Optimizer & Adam & Adam \\
        Adam \((\beta_1,\beta_2)\) & \((0.9,0.95)\) & \((0.9,0.95)\) \\
        Language / vision learning rate &
        \(4\times10^{-5}\) / \(7\times10^{-6}\) &
        \(2\times10^{-5}\) / \(7\times10^{-6}\) \\
        Warmup & 5\% of scheduled steps & 5\% of scheduled steps \\
        Learning rate after warmup & Constant & Cosine decay \\
        Weight decay / gradient clipping & 0.1 / 1.0 & 0.1 / 1.0 \\
        Global batch size & 1,024 & 128 \\
        Configured schedule & 3 epochs; 11,127 steps & 3 epochs; 79,404 steps \\
        Maximum sequence length & 32,768 tokens & 32,768 tokens \\
        Visual context & Full frame & Full frame and \(512\times384\) minimap \\
        Interaction history & 20 rounds & Up to 16 rounds, retain 8 \\
        Model precision & BF16 & BF16 \\
        \bottomrule
    \end{tabular}
    \caption{Two-stage Policy training configuration for the data in
    Section~\ref{sec:data_sources}. The action-pretraining schedule spans
    11,127 steps; instruction tuning uses its step-3,500 checkpoint.
    Instruction tuning has 26,468 steps per epoch and 3,970 warmup steps,
    and applies the repeated-action weighting in
    Section~\ref{sec:training_details}.}
    \label{tab:recipe}
\end{table}

Global batch size counts packed training windows.

\FloatBarrier

\section{Infrastructure Ablations and Historical Window Protocols}
\label{app:infra_details}

\subsection{Control Admission and Execution}
\label{app:control_scheduling}
Request admission and action-window handling follow
Section~\ref{sec:policy_scheduling}. The executor holds the last applied
controller state across handoffs; if fresh states cease to arrive, a watchdog
releases inputs and initiates a pause.

For upload and prefix preparation measured on the same clock, the residual
wait after upload is
\[
 W_{\mathrm{prep}}=\max\bigl(0,t_{\mathrm{prefix}}-t_{\mathrm{upload}}\bigr),
\]
where \(t_{\mathrm{prefix}}\) and \(t_{\mathrm{upload}}\) are their respective
completion times. Replay request latency ends at complete response delivery
and excludes subsequent waiting for scheduled native execution.

\subsection{History-Strategy Comparison}
\label{app:history_retest}
\begin{figure}[!hbp]
    \centering
    \includegraphics[width=\linewidth]{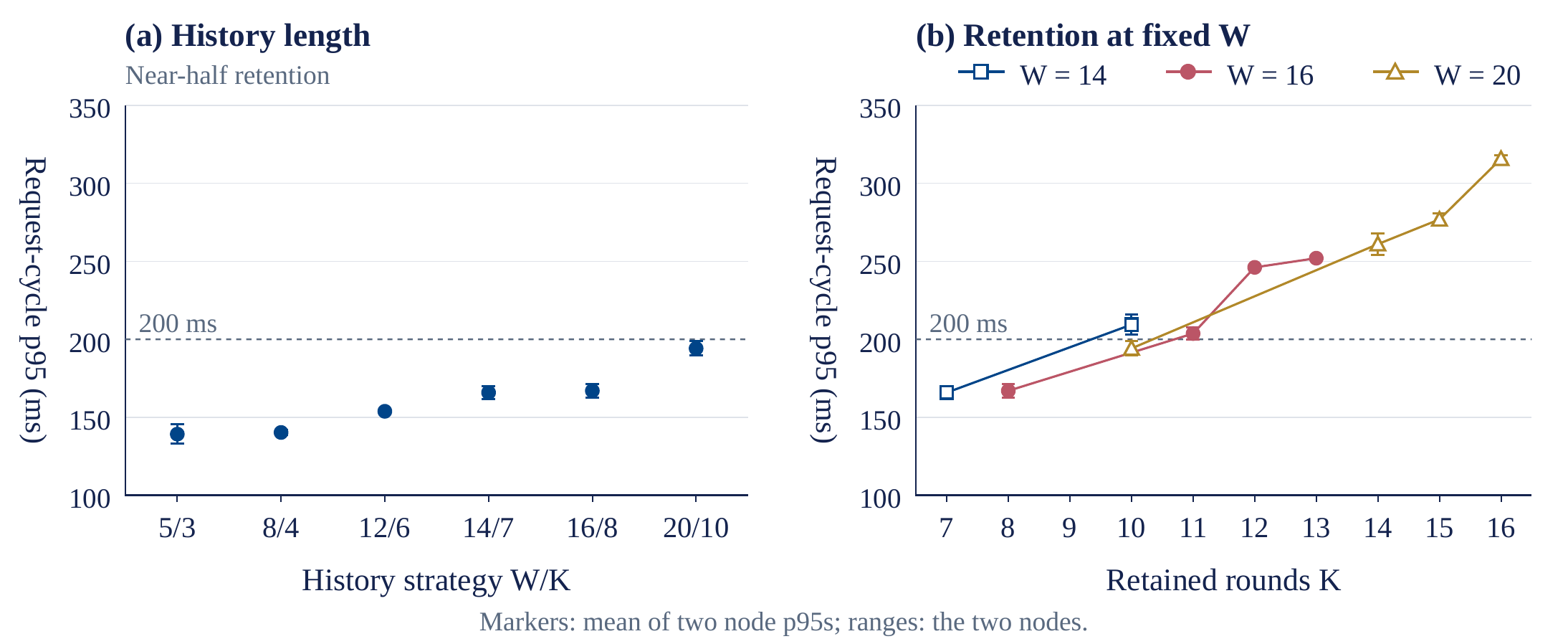}
    \caption{Request-cycle latency under different history strategies.
    (a) Window length with approximately half of the rounds retained.
    (b) Retained rounds \(K\) at fixed window lengths \(W\).
    Dashed lines mark the 200\,ms control interval.}
    \label{fig:history_practical}
\end{figure}

We compare 13 dual-route \((W,K)\) history strategies in terms of complete
request-cycle latency. Here, \(W\) is the maximum number of history rounds
and \(K\) is the number retained after trimming.
Figure~\ref{fig:history_practical} first varies window length while retaining
approximately half of the rounds, then varies retention at fixed
\(W\in\{14,16,20\}\). Together, the panels cover all 13 strategies.

The comparison uses a fixed visual-feature cache of 32,768 embeddings per
route. Two independent eight-H200 nodes each host two TP4 routes, with three
160-observation repetitions per node and setting. The configured prefill
and encoder-compute budgets remain 16,384, with a production graph ceiling
of 8192.

Each marker averages the two node-specific request-cycle p95s, and its
vertical range spans those two values. Each node's p95 pools three runs
after excluding the first \(W\) decisions of each run and includes all
generated responses.

At the default 16/8 setting, complete request-cycle p95 is
162.5--171.5\,ms across the two nodes, with eight decisions between trims.
Extending the window to 20/10 increases p95 to 189.5--198.8\,ms.
At fixed \(W=16\), retaining twelve rounds instead of eight increases
p95 to 245.3--246.9\,ms and halves the interval between trims.
We use 16/8 to balance retained history and latency margin relative to
the 200\,ms control interval.

Every session consumes the same ordered raw frames, with source publication
every 200\,ms and no frame skipping. The frozen client derives each minimap
from the full frame. The first 24 OCR blocks match the existing fixtures;
subsequent observations deliver no new OCR. Both routes are warmed on the
first already-known observation; KV, processor, and encoder caches reset
before each session. The replay retains full autonomous outputs, verifies
each prepared token/media prefix, and executes no external Brain calls or
game inputs. Admission backlog is recorded separately from request latency.

\paragraph{Encoder-cache diagnostic.}
The full experiment additionally includes four serial controls and repeats
all 17 settings with a 16,384-entry cache. The two nodes take opposite
cache-arm orders. Across both cache capacities, three repetitions of 160
observations per node/cache/setting give 204 sessions, 32,640 autonomous
decisions, and 29,688 steady requests after excluding the first \(W\)
decisions per session.

A separate fixed-input diagnostic uses six frozen 15-/16-round prompts
from three recording windows, exactly 30 output tokens, and twelve formal
repetitions per prompt/cache/placement. Capacities exchange physical GPU
groups in a second placement. Text KV resets before every request; encoder
and processor caches persist after two warmups. All 288 formal requests
match their frozen-loader input digests. Allocation counters are enabled
equally in this diagnostic and disabled in the replay and request benchmarks.
Fifteen distinct rounds fit in a 16,384-entry encoder cache; sixteen
require 17,152 embeddings. In the fixed-input sixteen-round control,
increasing capacity to 32,768 reduces mean first-token latency from
440.0\,ms to 261.0\,ms and eliminates new visual-embedding allocations
after warmup. At 20/16, the same capacity increase reduces complete-cycle
p95 from 455.7--488.8\,ms to 313.6--317.8\,ms across the two nodes.
Autonomous replay trajectories can diverge across configurations; their
prompt/output identity is retained separately from the fixed-input control.
Encoder storage, processor cache, and text KV/recurrent storage
are separate budgets; configured batch tokens alone do not determine actual
encoder capacity because the runtime also accommodates maximum-sized items.

\subsection{Controlled Inference Benchmark Protocol}
\label{app:infra_budget_controls}
The fixed-window suite evaluates Qwen3.5-9B in BF16 on two independent
8-H200 nodes. Every configuration receives the same 16 source-video windows,
with 15 completed history rounds and the current observation. Windows are
selected before timing by a fixed hash order over distinct videos. Each
round contains a full frame and minimap, contributing 880 and 192 visual
embeddings. The 32,768-embedding cache covers all 17,152 embeddings in the
16-round input; the batch and encoder-compute budgets remain 16,384 tokens.
Each scene is repeated five times per configuration and node, yielding
80 requests per node. The two nodes run configurations in opposite orders.

Every configuration first passes 80 separate autonomous-generation acceptance
cases, including empty reasoning, small updates, and Brain calls. Timed
requests then generate exactly 20 tokens, with matched input tokens and
ordered images. This fixed-length benchmark is separate from natural stopping
at the end of an action. Image upload and prior prefix preparation are outside
the timer, which ends at complete response delivery. Preparation uses the
normal prefill API with one disposable completion token; that token never
enters dialogue history. All reported means average the two equally sized
node cohorts. TP scaling uses a common capture ceiling of 512 tokens;
component removals use TP4 with an 8192-token ceiling. The deployed TP4
routes use the latter configuration and the same 32,768-embedding cache.

\paragraph{State-refresh handoff.}
\label{app:infra_handoff}
We compare stopping and resetting the active node, preparing a fresh session
in the background on the same node, and preparing it on a second node.
Each node hosts two TP4 Policy routes with identical Qwen3.5-9B BF16 weights,
an 8192-token graph-capture limit, and a 32,768-embedding visual cache.
Two recorded state-update scenarios, two active-node assignments, and six
repetitions give 24 trials per strategy; all 72 handoffs passed the readiness
checks. The current screenshot is fixed within each trial and matched across
strategies; the old session starts from eight recorded rounds under the
16/8 policy. Responses are generated autonomously. Updates alternate between
two phases of the 200\,ms observation grid.
Timing starts when the revised state becomes available, excluding Brain
computation. Handoff readiness requires a complete valid response and its
continuation prefix. The availability gap is the largest interval between
eligible action states during each transition, measured at the replay client;
it excludes physical controller execution.
Table~\ref{tab:infra_handoff} shows that a second node improves continuity
and avoids same-node contention. Immediate reset reaches the revised session
sooner on average, but produces a larger availability gap.
\input{tables/infra_handoff}

\subsection{Execution and Serving Strategy Controls}
\input{tables/infra_policy_efficiency}
Table~\ref{tab:infra_policy_efficiency}(a) removes six components independently
from the full TP4 service. All conditions use the shared full-window protocol
above and prepare history before the timed request. Generic input preprocessing
uses one processor thread and the original token validation and placeholder
processing; the prompt-assembly control changes only the image-token processing
implementation. Eager execution disables compiled execution and graph replay
together, so its latency does not isolate CUDA Graphs alone. The per-component
increases are not additive.

\begin{figure}[!htbp]
    \centering
    \includegraphics[width=\linewidth]{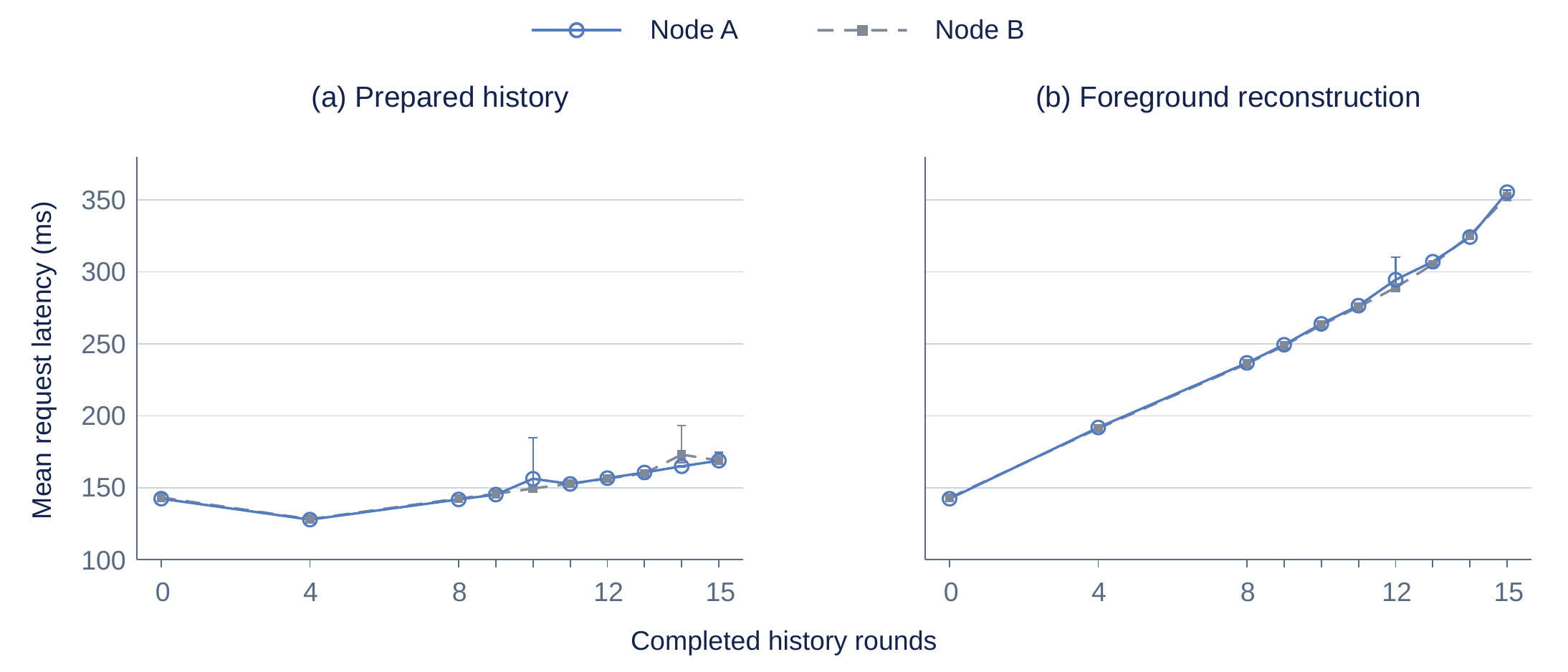}
    \caption{TP4 request latency as retained history grows, for 20-token responses, with the
    current observation fixed within each scene. History is prepared before timing in (a).
    In (b), text state is rebuilt inside the request using cached visual features. At 15
    completed rounds, the means across the two nodes are 168.9 and 353.8\,ms,
    respectively. Each point averages all 80 measured requests on one node (16 scenes,
    five repetitions each). Whiskers span the minimum and maximum of the five repeat
    means. Both panels use an 8192-token capture limit and a 32,768-embedding visual
    cache.}
    \label{fig:infra_history_length}
\end{figure}

\paragraph{Latency versus retained history length.}
\label{app:history_length}
Figure~\ref{fig:infra_history_length} uses the full TP4 service and the same
16 scenes at 0, 4, and 8--15 completed history rounds. Within each scene,
the current observation and persistent context stay fixed; shorter inputs
retain the most recent history suffix. Each point has five repetitions per
scene and node, with exactly 20 output tokens. Prepared requests reuse a
verified prefix. Reconstruction requests discard text KV after preparation
while retaining the same history visual features, so rebuilding text state
falls inside the timer. The 15-round prepared samples are shared with the
main component reference, not counted twice. This gives 4,960 distinct formal
requests across both nodes and all configurations.

With preparation, the mean rises from 142.2\,ms at eight history rounds to
168.9\,ms at fifteen; foreground reconstruction rises from 236.4 to
353.8\,ms. This suffix sweep measures request cost at a given history length.
Rolling-window trim frequency and dual-route overlap are measured separately
by the replay experiments.

\clearpage
\subsection{Live Control Cadence and End-to-End Latency}
\label{app:live_control_cadence}
We measure online control cadence at native action-packet starts.
Table~\ref{tab:live_control_cadence} summarizes a live run with a 16/8
observation-round window, where each round contains a full frame and
minimap. Across 1,023 executed packets, the control rate is close to
the 5\,Hz schedule.

\input{tables/live_control_cadence}

Action gaps measure the intervals between consecutive native packet starts.
Frame-to-action latency runs from the source-frame timestamp to the
corresponding native packet start, including processing and scheduling
delays. Frame capture and request preparation introduce a time offset in
the observation--action stream. Pipelined execution allows these stages to
overlap action execution, preserving approximately 200\,ms spacing between
successive actions. The measured frame-to-action p95 of 294.40\,ms is thus
compatible with near-5\,Hz control.

%% file: tables/infra_handoff.tex
\begin{table}[!htbp]
\centering
\small
\setlength{\tabcolsep}{5pt}
\begin{tabular}{lrrr}
\toprule
State-refresh strategy & Ready mean & Gap mean & Gap p95 \\
 & (ms) & (ms) & (ms) \\
\midrule
Single-node reset & 244.4 & 289.9 & 378.9 \\
Same-node background & 364.1 & 263.5 & 269.5 \\
Second-node background & 265.2 & 208.9 & 215.4 \\
\bottomrule
\end{tabular}
\caption{Controlled state-refresh replay, 24 trials per strategy.
Ready is elapsed time from state availability to handoff readiness.
Gap statistics summarize each trial's maximum gap between eligible action
states; the background strategies admit the candidate at full handoff
readiness, while immediate reset admits its first executable state.}
\label{tab:infra_handoff}
\end{table}

%% file: tables/infra_policy_efficiency.tex
\begin{table}[!htbp]
\centering\small
\setlength{\tabcolsep}{4pt}
\renewcommand{\arraystretch}{1.12}
\begin{tabular*}{\linewidth}{@{\extracolsep{\fill}}lrrrr@{}}
\toprule
\multicolumn{5}{l}{\textbf{(a) Independent component removals: TP4, capture 8192}} \\
\midrule
Configuration & Node A & Node B & Mean (ms) & Ratio \\
\midrule
\textbf{Full service} & 168.8 & 169.0 & \textbf{168.9} & 1.00$\times$ \\
Without compilation + graph replay & 883.2 & 875.6 & 879.4 & 5.21$\times$ \\
Without shared-memory image transfer & 573.4 & 576.6 & 575.0 & 3.40$\times$ \\
Without asynchronous scheduling & 219.3 & 217.0 & 218.2 & 1.29$\times$ \\
Without input preprocessing & 207.8 & 207.6 & 207.7 & 1.23$\times$ \\
Without vision-encoder graph replay & 185.0 & 184.0 & 184.5 & 1.09$\times$ \\
Without direct image-token assembly & 184.9 & 184.7 & 184.8 & 1.09$\times$ \\
\bottomrule
\end{tabular*}
\par\medskip
\begin{tabular*}{\linewidth}{@{\extracolsep{\fill}}lrrr@{}}
\toprule
\multicolumn{4}{l}{\textbf{(b) Baseline and TP scaling: shared capture 512}} \\
\midrule
Configuration & Node A & Node B & Mean (ms) \\
\midrule
Baseline TP1 & 622.9 & 618.9 & 620.9 \\
Optimized TP1 & 246.4 & 243.0 & 244.7 \\
Optimized TP2 & 195.9 & 197.3 & 196.6 \\
Optimized TP4 & 180.6 & 178.2 & 179.4 \\
Optimized TP8 & 177.9 & 179.6 & 178.7 \\
\bottomrule
\end{tabular*}
\caption{Fixed-window inference measurements for Qwen3.5-9B (BF16, H200, 20 generated tokens). Every configuration uses the same 16 scenes, each with 15 completed history rounds and the current observation, repeated five times per node. Both panels use a 32,768-embedding visual cache and a 16,384-token batch budget. Capture limits count tokens per execution. Component ratios use the full-service mean and are not additive. Upload and prior prefix preparation are excluded.}
\label{tab:infra_policy_efficiency}
\label{tab:infra_optimization_ablations}
\end{table}

%% file: tables/live_control_cadence.tex
\begin{table}[!htbp]
    \centering
    \small
    \setlength{\tabcolsep}{6pt}
    \begin{tabular}{lrrrr}
        \toprule
        Window & Actions & Active Hz & \shortstack{Action gap\\p95 (ms)} & \shortstack{Frame--action\\p95 (ms)} \\
        \midrule
        16/8 & 1023 & 4.979 & 200.98 & 294.40 \\
        \bottomrule
    \end{tabular}
    \caption{Live control cadence and end-to-end latency with a 16/8
    observation-round window. Actions count unique native packet starts;
    active Hz is their rate during active control. The run contains no
    control-phase pauses.}
    \label{tab:live_control_cadence}
\end{table}